%% file: jair_aa.tex
\documentclass[twoside,11pt]{article}
\usepackage{jair, theapa, rawfonts}
\usepackage[misc]{ifsym}
\usepackage{ulem}
\usepackage{soul}
\usepackage{url}
\usepackage{latexsym}
\usepackage[utf8]{inputenc}
\usepackage{multicol}
\usepackage{caption}
\usepackage{multirow}
\usepackage{graphicx}
\usepackage{amsmath}
\usepackage{booktabs}
\usepackage{algorithm}
\usepackage{algorithmic}
\usepackage{stfloats}
\usepackage{comment}
\usepackage{ulem}
\usepackage{color}
\usepackage{array}
\usepackage{booktabs}
\newcommand{\tabincell}[2]{\begin{tabular}{@{}#1@{}}#2\end{tabular}}
\ShortHeadings{Out of Context: A New Clue for Context Modeling of Aspect-based Sentiment Analysis}
{Xing, Tsang}

\begin{document}

\title{Out of Context: A New Clue for Context Modeling of Aspect-based Sentiment Analysis}

\author{\name Bowen Xing \email Bowen.Xing@student.uts.edu.au \\
       \addr Australian Artificial Intelligence Institute, University of Technology Sydney \\Ultimo, NSW 2007, Australia
       \AND
       \name Ivor W. Tsang \Letter \email Ivor.Tsang@uts.edu.au \\
       \addr Australian Artificial Intelligence Institute, University of Technology Sydney \\Ultimo, NSW 2007, Australia
       }


\maketitle
\input{abstract}
\input{introduction}
\input{Motivation}
\input{relatedwork}

\input{Aspect-aware}

\input{experiment}

\input{result}

\input{discussion}
\input{conclusion}

\input{acknowledge}

\vskip 0.2in
\bibliography{ref}
\bibliographystyle{theapa}

\end{document}

%% file: abstract.tex
\begin{abstract}
Aspect-based sentiment analysis (ABSA) aims to predict the sentiment expressed in a review with respect to a given aspect. 
The core of ABSA is to model the interaction between the context and given aspect to extract  aspect-related information.
In prior work, attention mechanisms and dependency graph networks are commonly adopted to capture the relations between the context and given aspect. 
And the weighted sum of context hidden states is used as the final representation fed to the classifier. 
However, the information related to the given aspect may be already discarded and adverse information may be retained in the context modeling processes of existing models.
Such a problem cannot be solved by subsequent modules due to two reasons. First, their operations are conducted on the encoder-generated context hidden states, whose value cannot be changed after the encoder. Second, existing encoders only consider the context while not the given aspect.
To address this problem, we argue the given aspect should be considered as a new clue out of context in the context modeling process.
As for solutions, we design three streams of aspect-aware context encoders: an aspect-aware LSTM, an aspect-aware GCN, and three aspect-aware BERTs. They are dedicated to generating aspect-aware hidden states which are tailored for the ABSA task.
In these aspect-aware context encoders, the semantics of the given aspect is used to regulate the information flow.
Consequently, the aspect-related information can be retained and aspect-irrelevant information can be excluded in the generated hidden states.
We conduct extensive experiments on several benchmark datasets with empirical analysis, demonstrating the efficacies and advantages of our proposed aspect-aware context encoders.

\end{abstract}

%% file: introduction.tex
\section{Introduction}\label{sec:introduction}
With increasing numbers of comments on the Internet, sentiment analysis has attracted interesting interest from both research and industry. 
Aspect-based sentiment analysis is a fundamental and challenging task in sentiment analysis, which aims to infer the sentiment expressed in sentences with respect to given aspects. 
For example, there is a review: \textit{``The salad is so delicious but the soup tastes bad."}, in which the opinion over the \textit{`salad'} is positive, while the opinion over the \textit{`soup'} is negative.
 In this case, aspects are explicitly included in the comments, and predicting aspect-based sentiment polarities of this kind of comments is termed as aspect term sentiment analysis (ATSA) or target sentiment analysis (TSA).
 There is another case where the aspect is mentioned but maybe not explicitly included in the comment.
 For example, \textit{``Although the dinner is expensive, waiters are so warm-hearted!"}. We can observe that there are two aspects mentioned in this comment: \textit{price} and \textit{service} with completely opposite sentiment polarities.
 Predicting aspect-based sentiment polarities of this kind of comments is termed as aspect category sentiment analysis (ACSA), and the aspect categories belong to a predefined set.
 In this paper, we collectively refer to aspect term and aspect category as aspect. And our goal is aspect-based sentiment analysis (ABSA) including ATSA and ACSA, which are both classification tasks.

Long Short-Term Memory networks (LSTM) \cite{LSTM} is the most widely-used context encoder in ABSA task.
Previous models using LSTM as context encoders can be roughly divided into four categories:
 (1) The first category models conduct joint modeling of concatenated contexts and aspects. Attention-based LSTM with aspect embedding (ATAE-LSTM) \cite{ATAE} and modeling inter-aspect dependencies by LSTM (IAD-LSTM) \cite{IAD} model the context and aspect simultaneously via concatenating the aspect vector to each context word embedding in the embedding layer before LSTM.
 (2) In the secondary category, the models only use context words as input when modeling the context. Interactive attention networks (IAN) \cite{IAN} and aspect fusion LSTM (AF-LSTM)" \cite{Fusion} model the context alone while utilizing the aspect to study the interaction between context and aspect in the attention mechanism.
 (3) The methods in the third category additionally multiply a relative position weight to highlight the potential aspect-related context words. Recurrent attention network on memory (RAM) \cite{Tencent} assigns relative position weights to context hidden states before the attention mechanism.
 (4) The fourth category models are more special. After LSTM layer, they employ graph convolutional networks (GCN) or graph attention networks (GAT) to leverage the syntactic information by encoding the context's syntax graph. ASGCN \cite{asgcn} utilizes GCN to enhance the hidden states with the syntactical connections.
 Although recent LSTM-based models may adopt different subsequent modules from the above models, their context modeling process falls into the above four categories.

Bidirectional Encoder Representations from Transformer (BERT) is of a multi-layer bidirectional Transformers \cite{transformer} architecture.
BERT is a representative pre-trained language model.
Recently, it has achieved state-of-the-art results on heterogeneous NLP tasks \cite{s-bert,distillbert,bert-attack}, including ABSA task \cite{DGEDT,RGAT}.
BERT has two encoding manners: single sentence (SS) and sentence pair (SP), whose formal inputs are as follows:

{SS:} \texttt{[CLS] $s^1$, ..., $s^n$ [SEP]}

{SP:} \texttt{[CLS] $s_1^1$, ..., $s_1^{n_1}$ [SEP] $s_2^1$, ..., $s_2^{n_2}$ [SEP]}\\
where $s_1$ and $s_2$ denote the two sentences in SP modeling.
For convenience, in this paper, we refer the BERT of SS manner as BERT0 and the BERT of SP manner as BERT1.

LSTM, GCN, and BERT can be seen as the sequence-based context encoder, graph-based context encoder, and pre-trained context encoder, respectively.
All of them can generate or modify the inner values of the hidden states, which cannot be achieved by subsequent modules such as aspect-specific attention mechanisms \cite{IAN,asgcn}.
These context encoders are widely adopted in existing ABSA models and their generated hidden states are taken as input of subsequent modules.
However, there is a question which has never been considered: Are these hidden states good enough?

We argue that the aspect-related information may be discarded and the aspect-irrelevant information may be retained in the hidden states generated by LSTM, GCN and BERT.
The reason is that there is no aspect information introduced into the context modeling process of these context encoders, which cannot process the latent semantic space according to the aspect of current sample.
In this paper, we term this problem as the aspect-agnostic problem in the context modeling process.
And in Section 2 we depict this problem in details.

To solve the aspect-agnostic problem, we argue that the semantics of the given aspect should be explicitly introduced into context modeling process as a new clue out of the context. With the consideration of the given aspect, the aspect-aware context encoder can specifically retain useful information and eliminate aspect-irrelevant information in the generated aspect-aware hidden states which can improve ABSA.
Specifically, we propose three streams of aspect-aware context encoders which are based on LSTM, GCN and BERT, respectively.
Based on LSTM, we design an aspect-aware (AA) LSTM which augments vanilla LSTM with a novel aspect-aware (AA) mechanism.
The AA mechanism includes three aspect gates, which are corresponding to the input gate, forget gate, and output gate of LSTM cell respectively.
The aspect gates take the aspect vector and previous hidden state as input, producing a gate vector.
The gate vector controls the fraction of disturbance from the given aspect added to the internal value of the three LSTM gates.
In this way, when modeling the context, AALSTM can dynamically identify the aspect-related information as well as the aspect-irrelevant information.
With the consideration of the given aspect, AALSTM can retain the useful information in generated hidden states and prevent the harmful information from fusing into generated hidden states.
We propose the aspect-aware GCN by augmenting vanilla GCN with an aspect-aware convolution gate, which controls what and how much information from the neighbor nodes should be passed to the current node, regarding the specific aspect.
As for BERT, we do not change its internal network architecture so as to preserve its strong language modeling capability.
Skillfully utilizing the setting of the segment embedding and \texttt{[SEP]} token, we modify the input format of BERT to make BERT capture the aspect-aware \textbf{intra-sentence} dependencies when modeling the context.
We propose three aspect-aware (AA) BERT variants whose differences lie in the input formats.
Note that as AABERTs share the same parameters with standard BERT, they have no extra computation cost.

A preliminary version of this work has been presented in the conference paper \cite{AA}. The contribution of the previous version can be summarized as follows:
\begin{itemize}
    \item We discover the aspect-agnostic problem in ABSA task. To our knowledge, this is the first time that this problem is identified. To solve this problem, we propose a novel LSTM variant termed aspect-aware LSTM (AALSTM) to introduce the aspect into the process of context modeling.
    \item Considering that the aspect is the core information in this task, we fully exploit its potential by introducing it into the LSTM cells. We design three aspect gates to introduce the aspect into the input gate, forget gate and output gate in the LSTM cell. AALSTM can utilize aspect to improve the information flow and then generate more effective aspect-specific context hidden states tailored for ABSA task.
    \item We apply our proposed AALSTM to several representative LSTM-based models, and the experimental results on the benchmark datasets demonstrate the efficacy and generalization of our proposed AALSTM.
\end{itemize}
In this paper, we significantly extend our work from the previous version in the following aspects:
\begin{itemize}
    \item We discover that although GCN and BERT are widely adopted and have achieved promising performance in ABSA task, they also suffer the aspect-agnostic problem. 
    \item To solve the aspect-agnostic problem in GCN, we propose the aspect-aware GCN (AAGCN) by augmenting vanilla GCN with a novel aspect-aware convolution gate to introduce aspect semantics into the graph convolution process. 
    \item To solve the problem in BERT, we propose three aspect-aware BERT (AABERT) variants by skillfully modifying the input format of BERT. In this way, our AABERTs can model the intra-sentence dependency between the aspect and context words in a more appropriate way.
    \item We conduct extensive experiments to evaluate the proposed (Bi-)AALSTM, AAGCN and AABERTs on ABSA task. The results demonstrate that AABERTs can overpass vanilla BERT not only as a single model but also as the context encoder, and AAGCN can work well with both (Bi-)AALSTM and AABERTs. Equipped with our aspect-aware context encoders, the baselines proposed several years ago can beat up-to-date models, achieving new state-of-the-art performances.
\end{itemize}

The remainder of this paper is organized as follows. Section 2 depicts the details of the aspect-agnostic problem. Section 3 summaries the recent studies on ABSA task; Section 4 elaborates the details of our proposed AALSTM and AABERTs; Section 5 introduces the details of experiments; Section 6 gives the evaluation results and analysis; Section 7 discusses the proposed aspect-aware encoders and further investigate their properties and advantages; Section 8 gives the conclusion of this work.

%% file: Motivation.tex
\section{Aspect-agnostic Problem} \label{sec:problem}
When modeling the context, LSTM cells are aspect-agnostic because no aspect information is introduced to the cell to guidance the information flow.
Consequently, the generated hidden states contain the semantic information that is important to the \textbf{whole review} rather than the \textbf{given aspect}.
This is because LSTM inherently tends to retain the important information for the overall semantics of the whole review in generated hidden states.
However, considering the characters of ABSA task, a context word is valuable only if its semantics is helpful for predicting the sentiment of the given aspect.
On the contrary, if the information of a context word is aspect-irrelevant, it may be noise information and harmful for the prediction of aspect sentiment, and its semantics should be eliminated in the context modeling process of the context encoder.
LSTM cannot identify these two kinds of information when modeling the context for that no aspect information is considered in its cells.
As a result, the aspect-related information may be already discarded and adverse information may be retained in the hidden state generated by LSTM.

Specifically, the lack of the aspect information considered in LSTM cells may cause the following two issues.
 For a specific aspect, on one hand, some of the semantic information of the whole review context is useless.
 This aspect-irrelevant information would adversely harm the final context representation, especially in the situation where multiple aspects exist in one comment.
 This is because when LSTM encounters an important token for the overall sentence semantics, this token's information is retained in every follow-up hidden state.
 Consequently, even if perfect attention weights are produced by the attention mechanism, these hidden states also contain useless information in respect to the aspect.
 And the contained useless information is even magnified to some extent, for that the important tokens are assigned greater attention weights and these tokens may contain some useless information.
 On the other hand, the information important to the aspect may be not sufficiently kept in the hidden states because of their small contribution to the overall semantic information of the sentence.
We define the above issues as the \textit{aspect-agnostic problem} of LSTM in ABSA task. This is the first time this problem is discovered.
Concretely, we take two typical examples to illustrate the \textit{aspect-agnostic problem}.

\begin{figure*}[t]
 \centering
 \includegraphics[width = 1.0\textwidth]{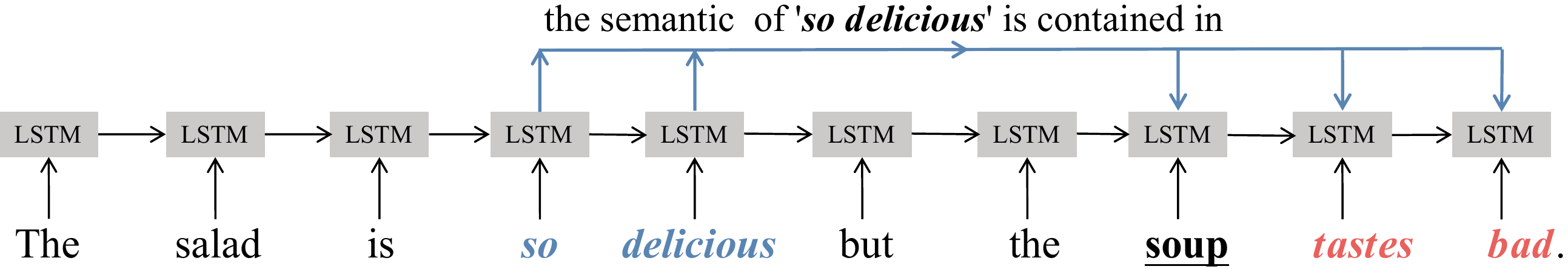}
 \caption{An example aiming to predict sentiment of soup}
 \label{fig: example1}
\end{figure*}

 The first example: \textit{``The salad is so delicious but the soup tastes bad."}, as shown in Fig.\ref{fig: example1}.
 There are two aspects (\textit{salad} and \textit{soup}) of opposite sentiment polarity.
 When inferring the sentiment polarity of \textit{soup}, the token `so delicious' which modifies \textit{salad} is also important to the sentence-level semantics of the whole review and therefore LSTM will retain its information in subsequent context words' hidden states, including the hidden states of \textit{tastes} and \textit{bad}.
Even if \textit{tastes} and \textit{bad} are assigned large attention weights by the attention mechanism, the semantics of `so delicious' will still be integrated into the final aspect-based context representation and enlarged by the large attention weights.
As a result, the adverse information from `so delicious' will harm the prediction of the aspect sentiment of \textit{soup}.
\begin{figure*}[t]
 \centering
 \includegraphics[width = 1.0\textwidth]{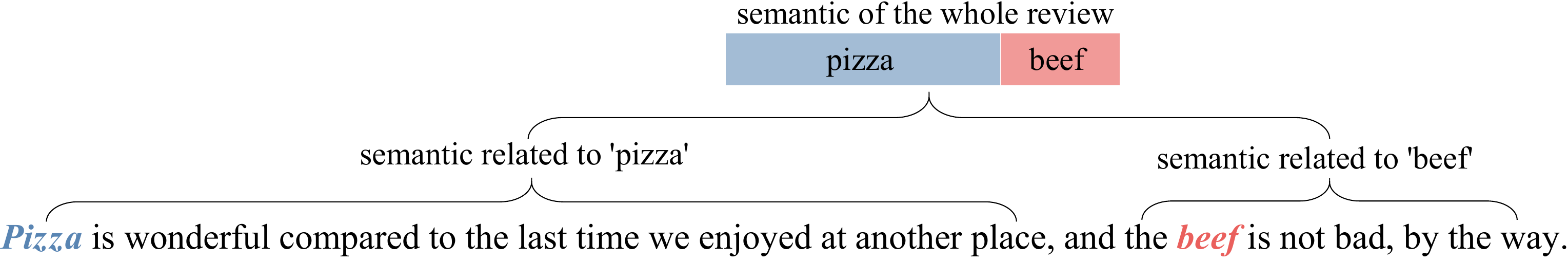}
 \caption{ An example aiming to predict sentiment of beef}
 \label{fig: example2}
\end{figure*}
 The second example: \textit{``Pizza is wonderful compared to the last time we enjoyed at another place, and the beef is not bad, by the way."}, as shown in Fig.\ref{fig: example2}. 
 We can find that this review is mainly about \textit{pizza} so LSTM cells will retain a lot of semantics that modifies \textit{pizza} while less semantics about \textit{beef} when modeling the context.
 When inferring the aspect sentiment of \textit{beef}, as LSTM is aspect-agnostic, probably some key information related to \textit{beef} is lost in the generated hidden states because of the relatively small contribution of \textit{beef}-related semantics to the overall semantics of the review. 

GCN are conceptual similar to LSTM as they both achieve the message passing from other words/nodes to current word/node.
In LSTM, the message passing is from previous words to current word, while in GCN the message is passed from neighbor nodes to current nodes.
And LSTM uses gate mechanisms to control the information flow, while in GCN this is achieved by graph convolutions.
In the graph convolution process, GCN does not know which nodes are aspect words nodes.
Then the important aspect-related information may be discarded and aspect-irrelevant information may be retained.
Hence, GCN suffers the same aspect-agnostic problem as LSTM.

\begin{figure*}[t]
 \centering
 \includegraphics[width = 1.0\textwidth]{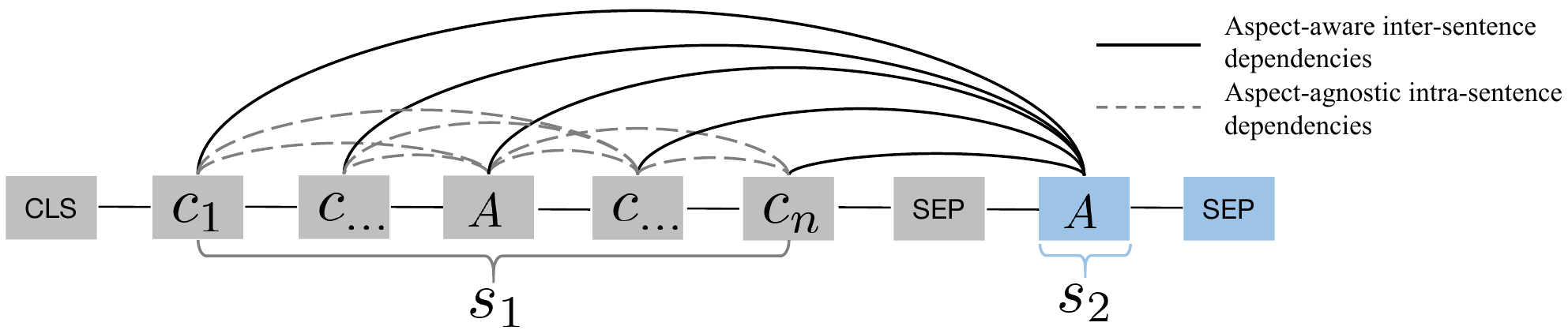}
 \caption{ Illustration of the sentence pair modeling process of BERT1. For simplify, some intra-sentence dependencies between context words are omitted.}
 \label{fig: bert1-framework}
\end{figure*}
As for BERT, BERT0 only takes the context as input and models the \textbf{intra-sentence} dependency without consideration of the given aspect. So it suffers the same aspect-agnostic problem as LSTM.
In BERT1, the context is in position $s_1$ and the aspect is in position $s_2$. 
As BERT1 models the concatenated context-aspect pair in the sentence-pair manner, it can extract aspect-related semantics from the context. Consequently, BERT1 is a strong baseline which is shown in Table \ref{bert-results}.
We attribute these improvements to BERT1's capability of capturing the inter-sentence dependencies of the context ($s_1$) and aspect $s_2$.
The sentence pair modeling process of BERT1 is illustrated in Fig \ref{fig: bert1-framework}.
We can observe that BERT1 regards the context ($s_1$) and the concatenated aspect ($s_2$)  as two individual sentences.
This is because that the separator token \texttt{[SEP]} and segment embeddings of BERT1 thoroughly separate the context ($s_1$) and aspect ($s_2$) in the latent space.
The inter-sentence dependencies captured by BERT1 is aspect-aware on account of the pre-training of the next sentence prediction task.
Although the sentence-pair modeling works like an attention mechanism considering the aspect in position $s_2$, there is not sufficient clue from the given aspect in the intra-sentence language modeling process of the context ($s_1$).
However, one of key characteristics of ABSA task is that the aspect is contained in the context.
As a result, the captured intra-sentence dependencies is general and aspect-agnostic, similar to the ones obtained by LSTM and BERT0.
This causes that some useful \textbf{intra-sentence} dependencies between the aspect and its related context words (especially the sentiment trigger words) may be lost in the context modeling process of BERT1.
Therefore, both BERT0 and BERT1 suffer the aspect-agnostic problem in ABSA task.

%% file: relatedwork.tex
\section{Related Work} \label{relatedwork}
Some traditional ABSA methods have achieved promising results, while they are labor-intensive as they focused on feature engineering or massive extra linguistic resources \cite{NRC2,DCU}. 
As deep learning achieved breakthrough success in representation learning \cite{dl-nature}, many recent ABSA approaches adopt deep neural networks to automatically extract features and generate the final aspect-based sentiment representation which is a dense vector fed into the classifier.
Since the attention mechanism was first introduced to the neural machine translation field \cite{NMT}, many sequence-based models utilize it to focus on the aspect sentiment trigger words for predicting the aspect's sentiment.
The attention mechanism in ABSA takes the aspect vector and the hidden states of context words as input.
Then it produces an attention vector to assign each context hidden state a weight according to its relevance to the given aspect.

The core of ABSA task is to model the interaction of the context and given aspect then extract the aspect-related semantics.
\cite{IAN} adopted two individual LSTM to model the context and aspect term.
The proposed interactive attention mechanism can learn the interaction between aspect and context, extracting the aspect-related information.
With the ability of extracting n-gram features, convolution neural network (CNN) \cite{textcnn} is applied to model the interaction between the context and aspect in some previous works \cite{gcae,pcnn}.
\cite{pcnn} utilized parameterized filters and parameterized gates to incorporate aspect information into CNN. As they declare, it was the first attempt using CNN to solve aspect-based sentiment analysis task.
There are also some memory networks \cite{end2endMN} (MNs)-based models \cite{DMN,Dm,tsmn}.
\cite{Dm} modeled dyadic interactions between aspect and context using neural tensor layers and associative layers with rich compositional operators.
\cite{tsmn} argues that for the case where several sentences are the same except for different targets, only relying on the attention mechanism is insufficient. It designed several memory networks having their own characters to solve the problem.
Capsule network \cite{capsule} is also exploited to tackle both sentiment analysis and aspect-based sentiment analysis tasks \cite{sacap,absacap,transcap}. To solve the problem of lacking labeled data of the ABSA task, \cite{transcap} transfers the knowledge of the document-level sentiment analysis task to aspect-based sentiment analysis. It designs an aspect routing approach and extends the vanilla dynamic routing approach by adapting it to the transfer learning framework. 

As dependency tree can shorten the distance between the aspect and its sentiment trigger words, graph neural network \cite{gnn} (GNN) such as graph convolutional network \cite{gcn,asgcn,DGEDT} and graph attention network (GAT) \cite{gat,graphatt,DGEDT} have been adopted for dependency graph modeling. \cite{asgcn,DGEDT,tgcn} utilized GCN to capture the inner connection between the given aspect and its sentiment trigger words on the semantic dependency tree.
\cite{RGAT} proposed a relational graph attention network that operates on the aspect-oriented pruned dependency tree.
On the pruned dependency tree, each context word is directly connected to the aspect with a defined relation.
In order to obtain more comprehensive and sufficient syntactic information, \cite{kagrmn,dignet} marry the local vanilla dependency graph and the global relation graph to let them compensate for each other.

As for the context encoder, LSTM is the most widely-used context encoder because of its advantages for sequence modeling \cite{TDLSTM,IAN,IAD,ATAE,Fusion,zhangyue,aaai2017,hrd}. 
And more recently, BERT has proved its power on heterogeneous NLP tasks \cite{s-bert,distillbert,bert-attack} and some ABSA models \cite{graphatt,RGAT,DGEDT,kvmn-eacl,tgcn} adopt BERT as context encoder to obtain high-quality hidden states of context words. 
\cite{bert_post} post-trained the BERT encoder on external document-level sentiment classification corpus to enhance BERT's ability of understanding sentiment semantics.
And \cite{chinese-oriented,ICPR} conducted adversarial training to improve aspect sentiment reasoning.

Although previous works can improve ABSA task by tackling different issues, the effectiveness of the context modeling processes of LSTM, GCN and BERT for ABSA task has never been inspected.
In this work, we discover the aspect-agnostic problem which is widely suffered by the context encoders of existing works.
And we propose the aspect-aware encoders to tackle this problem, improving ABSA from a new perspective.

%% file: Aspect-aware.tex
\section{Aspect-aware Context Encoders} \label{aa}
\begin{figure*}[h]
 \centering
 \includegraphics[width = 1.0\textwidth]{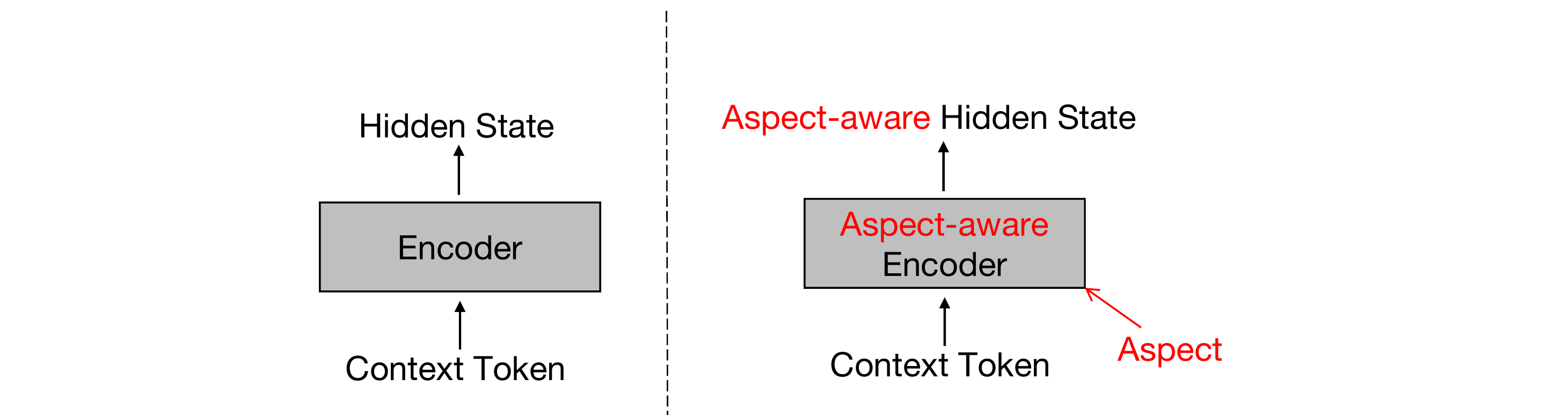}
 \caption{Conceptual illustrations of general context encoder and aspect-aware context encoder.
  Left picture illustrates general context encoder. Right picture illustrate our proposed aspect-aware context encoder.}
 \label{fig: conceptual}
\end{figure*}
Motivated by the observation and analysis of the aspect-agnostic problem, we propose to introduce explicit aspect semantics into the context encoder to make the context modeling process aspect-aware, as shown in Fig \ref{fig: conceptual}.
In this paper, we propose two streams of aspect-aware context encoders as solutions, whose backbones are vanilla LSTM and BERT respectively.
In the following sections, we will introduce the details of the proposed aspect-aware context encoders.

\subsection{Aspect-Aware LSTM}
\begin{figure}[t]
 \centering
 \includegraphics[scale = 0.5]{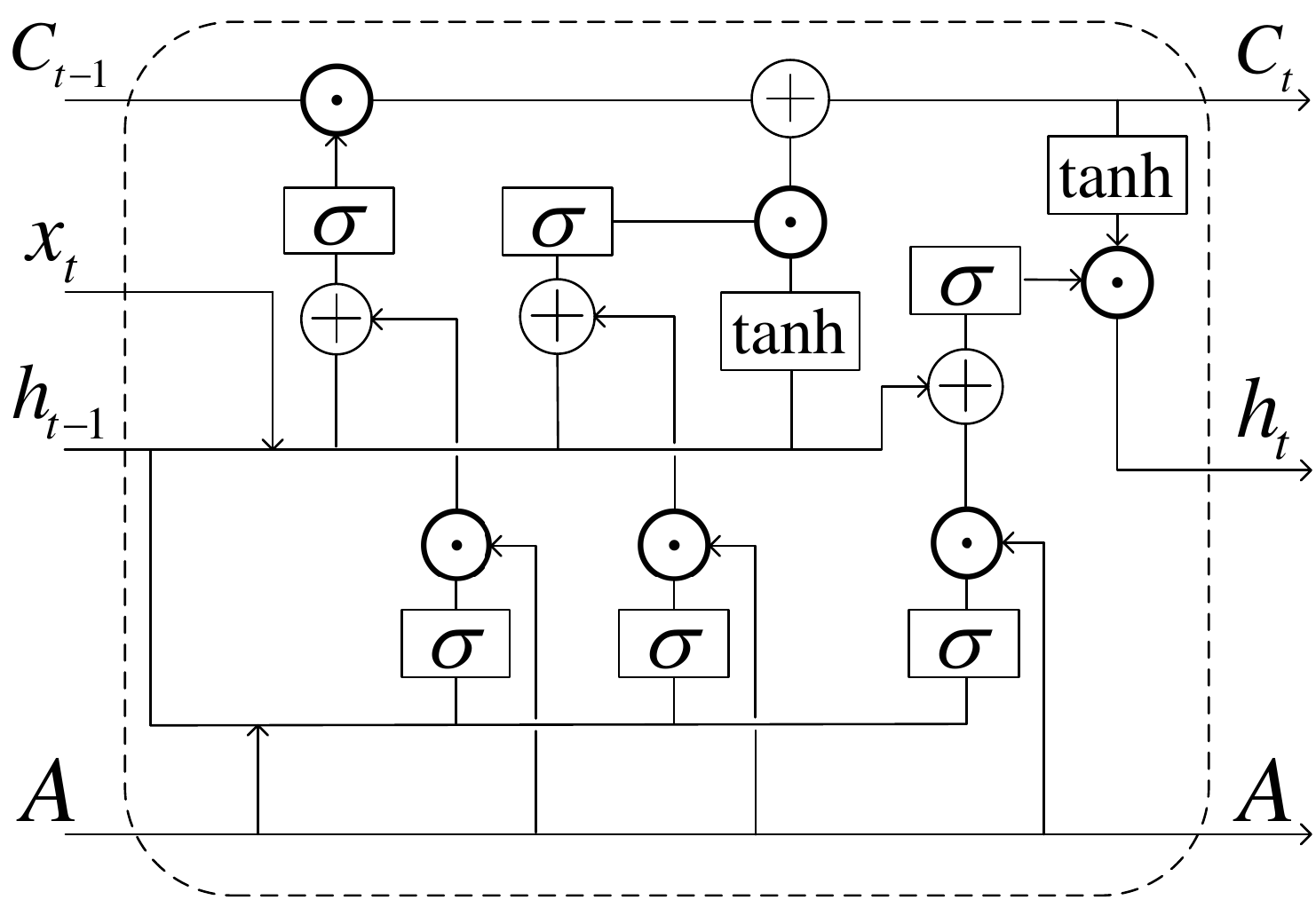}
 \caption{The overall architecture of AALSTM network}
 \label{lstm}
\end{figure}

Vanilla LSTM utilizes three gates (input gate, forget gate, and output gate) to model the dependency within the input word sequence and retain the important long dependency in the forward context modeling process.
We argue that the information of the given aspect should be introduced into LSTM cells to help regulate the information flow. 
Additionally, it is intuitive that in every time step the degree that the semantics of the given aspect is integrated into the three gates of classic LSTM should be dynamically adjusted according to the aspect information and the current semantic states.
Therefore, we design three aspect gates that control how much the aspect vector is integrated into the input gate, forget gate, and output gate respectively.
The aspect gate mechanism takes the previous hidden state and the aspect vector as input.
In this way, AALSTM can dynamically optimize the information flow in LSTM cells according to the given aspect, then generate effective and beneficial aspect-specific hidden states for ABSA task. 
Figure \ref{lstm} illustrates the overall architecture of our proposed AALSTM, which can be formalized as follows:
\begin{align}
a_i &= \sigma(W_{ai}\,[A, h_{t-1}] + b_{ai}) \\
I_t &= \sigma(W_I\,[x_t, h_{t-1}] + a_i \odot A + b_I) \\
a_f &= \sigma(W_{af}\,[A, h_{t-1}] + b_{af}) \\
f_t &= \sigma(W_f\,[x_t, h_{t-1}] + a_f \odot A + b_f) \\
\widetilde{C_t} &= tanh(W_C\,[x_i, h_{t-1}] + b_C) \\ \label{equo:candidate}
C_t &= f_t \odot C_{t-1} + I_t \odot \widetilde{C_t} \\ \label{equo:cell state}
a_o &= \sigma(W_{ao}\,[A, h_{t-1}] + b_{ao}) \\
o_t &= \sigma(W_o\,[x_t, h_{t-1}] + a_o \odot A + b_o) \\
h_t &= o_t * tanh({C_t}) 
\end{align}
where $x_t$ denotes current input context word embedding, $A$ is the aspect vector, $h_{t-1}$ is the previous hidden state, $h_t$ is the hidden state of current time step, $\sigma$ and $tanh$ are sigmoid and hyperbolic tangent functions, $\odot$ stands for element-wise multiplication, $W_{ai}, W_{af}, W_{ao}\in R^{da\times (dc+da)}$ and $W_I, W_f, W_C, W_o\in R^{dc\times 2dc}$ are weight matrices, $b_{ai}, b_{af}, b_{ao} \in R^{da}$, $b_I, b_f, b_C,\\ b_o \in R^{dc}$ are biases and $da, dc$ stand for the aspect vector's dimension and the number of hidden cells at AALSTM respectively.
$I_t, f_t, o_t \in R^{dc}$ stand for the input gate, forget gate, and output gate respectively.
The input gate controls the extent of updating the information from the current input.
The forget gate determines what information the current cell state should inherit from the previous cell state.
The output gate controls what information in the current cell state should be output as the hidden state of the current time step.
$a_i, a_f, a_o \in R^{da}$ stand for the aspect-input gate, aspect-forget gate and aspect-output gate respectively.
They determine the extent of integrating the aspect information into the input gate, forget gate, and output gate respectively.

AALSTM takes two strands of inputs: context word embeddings and the aspect vector.
At each time step, the context word entering the AALSTM dynamically varies according to the sequence of words in the sentence, while the aspect vector is identical.
Specifically, the aspect vector is the target representation in ATSA, and it is the aspect embedding in ACSA.
Next, we describe the different components of our proposed AALSTM in detail.
\subsubsection{Input Gates}
The input gate $I_t$ controls how much new information from the input context word embedding can be transferred into the cell state.
While the aspect-input gate $a_i$ controls how much the aspect information should be integrated into the input gate $I_t$.
The difference between $I_t$ in AALSTM and vanilla LSTM lies in the weighted aspect vector input into $I_t$.
The aspect-input gate $a_i$ is computed by $h_{t-1}$ and $A$ (Eq. 1).
$h_{t-1}$ can be regarded as the previous semantic representation of the partial sentence which has been processed in the past time steps.
Hence, the aspect-input gate $a_i$ is controlled by the previous semantic representation and the aspect vector $A$.
In $I_t$, the dynamically weighted aspect information $a_i \odot A$ is added to the original internal value calculated by $x_t$ and $h_{t-1}$ (Eq. 2).
Thereby, the dynamically adjusted disturbance from the given aspect can guide $I_t$ to determine what information from the current input context word embedding should be transferred into the cell state.

\subsubsection{Forget Gates}
The forget gate $f_t$ abandons trivial information and retains key information from previous cell state $C_{t-1}$.
The aspect-input gate $a_f$ controls how much the aspect vector should be integrated into the forget gate $f_t$.
The difference between AALSTM and vanilla LSTM in $f_t$ is the introduction of weighted aspect vector.
And the aspect-forget gate $a_f$ is computed by $h_{t-1}$ and $A$ (Eq. 3).
Therefore, the extent of the integration of aspect information into $f_t$ is decided by the previous semantic representation and the aspect vector $A$.
In $f_t$, $a_f \odot A$ is added to the original internal value calculated by $x_t$ and $h_{t-1}$ (Eq. 4).
Thereby, the dynamically adjusted disturbance from the given aspect information can guide $f_t$ to select aspect-related information from the previous cell state and retain it in the current cell state.
In the meantime, aspect-irrelevant information is abandoned.
\subsubsection{Candidate Cell and Current Cell}
The candidate cell $\widetilde{C_t}$ represents the alternative input content.
The current cell $C_t$ updates its state by selecting important information from previous cell state $C_{t-1}$ and the candidate cell $\widetilde{C_t}$.
From Eq. 5 we can observe that there is two kinds of information contained in the alternative input content $\widetilde{C_t}$: the last hidden state $h_{t-1}$ and current input context embedding $x_{t}$.
While the information in current cell state ${C_t}$ consists of the information from previous cell state $C_{t-1}$ and candidate cell $\widetilde{C_t}$, as shown in Eq. 6.
Considering that the information in $h_{t-1}$ comes from previous cell state $C_{t-1}$, the only source of the information contained in cell state ${C_t}$ and hidden state $h_t$ is the input context word embeddings.
So our proposed AALSTM only leverages the given aspect information to regulate the information flow in LSTM cells instead of fusing the aspect information into cell state nor hidden states.

\subsubsection{Output Gates}
The output gate $o_t$ controls what information of the current cell state should be output as the hidden state of the current context word.
The aspect-output gate $a_o$ controls what fraction of aspect should be integrated into the output gate $o_t$.
The difference between our proposed AALSTM and the vanilla LSTM in $o_t$ is the integration of the weighted aspect vector into $o_t$.
And the aspect-output gate $a_o$ is computed by $h_{t-1}$ and $A$ (Eq. 7).
Therefore, the degree of how much the aspect information is integrated into $o_t$ is decided by the previous semantic representation and the aspect vector $A$.
In $f_t$, the dynamically weighted aspect information vector $a_o \odot A$ is added to the original internal value calculated by $x_t$ and $h_{t-1}$ (Eq. 8).
Thereby, the optimized disturbance from the given aspect information can guide $o_t$ to specifically select the appropriate information from the current cell state as the hidden state of the current input context word.

\subsection{Aspect-aware Graph Convolutional Networks}
In ABSA, the function of GCN is encoding the local syntactic connections represented by the adjacent matrix derived from the syntax graph predicted by the off-the-shelf dependency parser.
The graph convolution operation of vanilla GCN can be written as:
\begin{equation}
h_i^l= \sum_{j\in \mathcal{N}_i} {W_g} \ h_j^{l-1} /(d_i + 1) + b_g\\
\end{equation}
where $l$ denotes $l-$th GCN layer, $\mathcal{N}_i$ denotes the set of $i-$node's neighbor nodes, $d_i$ is the degree of $i-$node in the syntax tree, $W_g$ is weight matrix and $b_g$ is bias.
\begin{figure}[ht]
 \centering
 \includegraphics[width = 0.7\textwidth]{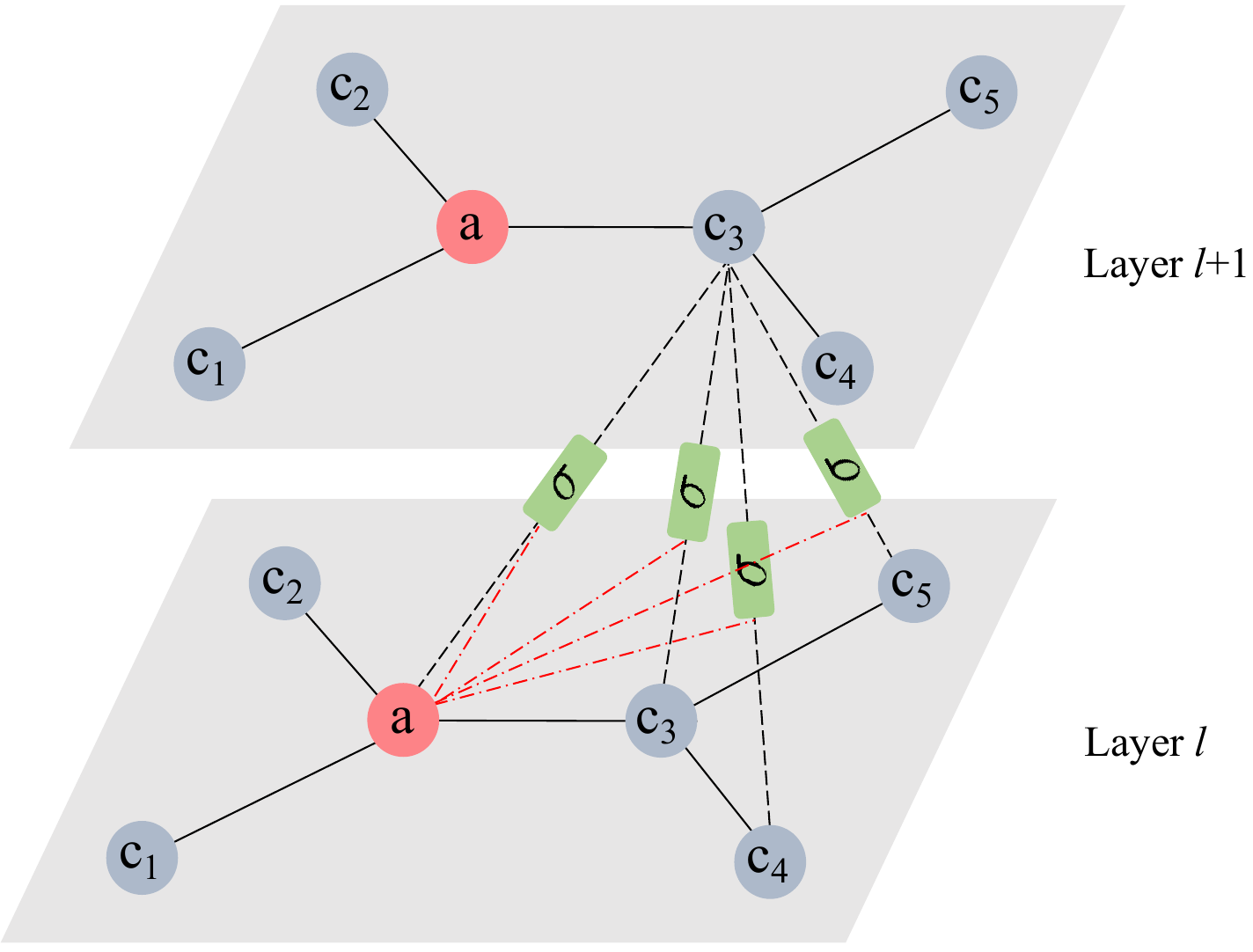}
 \caption{The architecture of our proposed aspect-aware GCN. The black dash line denotes the message passing from layer $l$ to layer $l+1$. The red dash line denotes the aspect semantics helps control what and how much information should be passed from layer $l$ to layer $l+1$.}
 \label{fig:aagcn}
\end{figure}

To tackle the aspect-agnostic problem of GCN, we propose the aspect-aware GCN (AAGCN) by augment vanilla GCN with the aspect-aware convolution gate.
The architecture of AAGCN is illustrated in Figure \ref{fig:aagcn}.
Our AAGCN can be formulated as:
\begin{align}
h_i^l=& \sum_{j\in \mathcal{N}_i} {W_g} \ h_j^{l-1}\odot a_c^j /(d_i + 1) + b_g\\
a_c^j  =& \sigma(W_{ac}\,[A, h_j^{l-1}] + b_{ac})
\end{align}
where $A$ denotes the aspect representation, $W_{ac}$ and $b_{ac}$ are the weight matrix and bias respectively.
The input of aspect-aware convolution gate is the aspect representation and the hidden state of the neighbor node.
Considering the semantics of the specific aspect and the hidden state of the neighbor node, the aspect-aware convolution gate output a vector which determines what and how much information from the neighbor node should be transfered into the current node.
In this way, AAGCN can aggregate the aspect-related information and eliminate aspect-irrelevant information harmful for ABSA in the process of aspect-aware graph convolution.
Then AAGCN can generate better hidden states which can improve the performances of ABSA.

\subsection{Aspect-aware BERT}
\begin{figure}[ht]
 \centering
 \includegraphics[width = 0.7\textwidth]{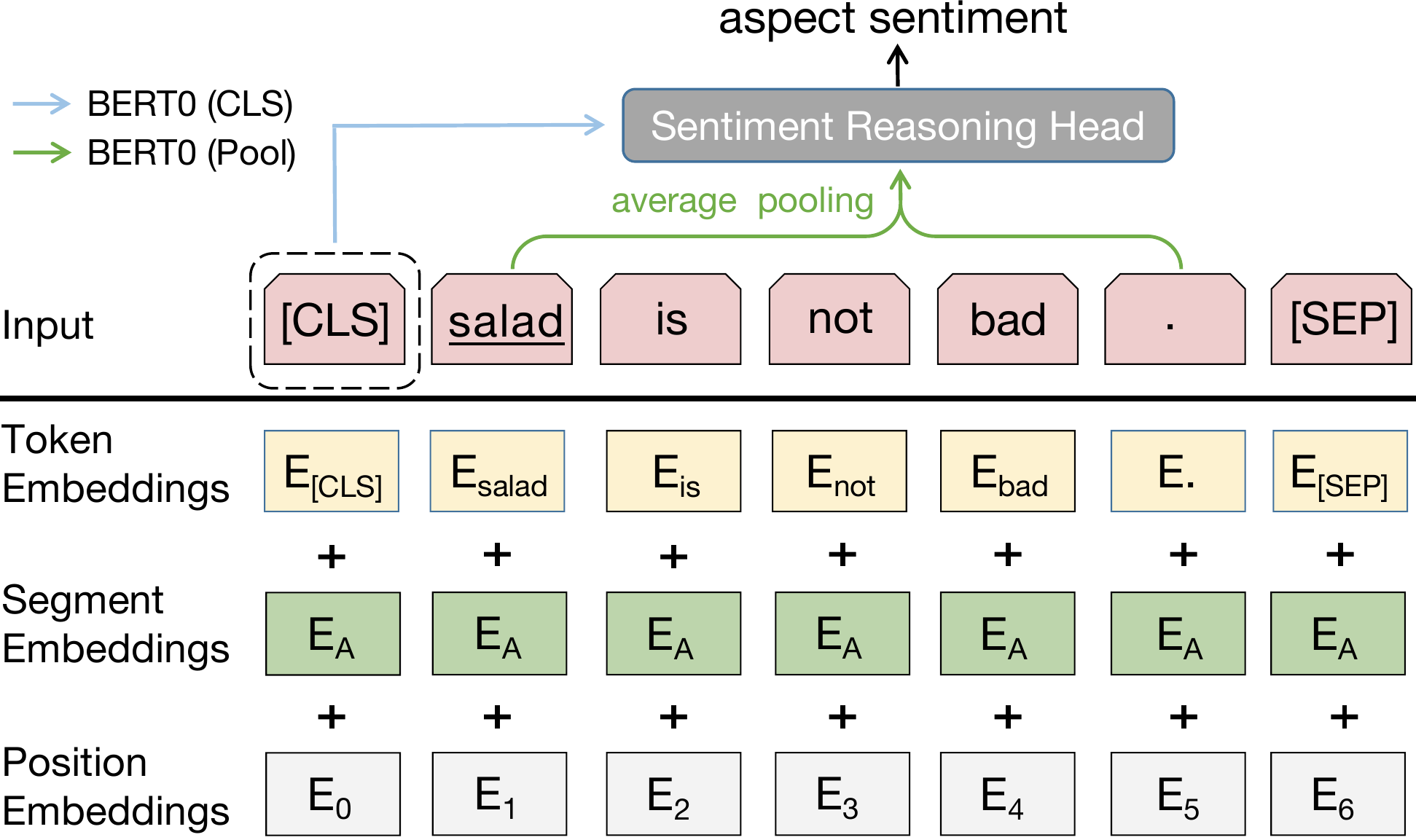}
 \caption{Illustration  of BERT0 (CLS) and BERT0 (Pool). }
 \label{fig:bert0}
\end{figure}
\begin{figure}[ht]
 \centering
 \includegraphics[width = 0.7\textwidth]{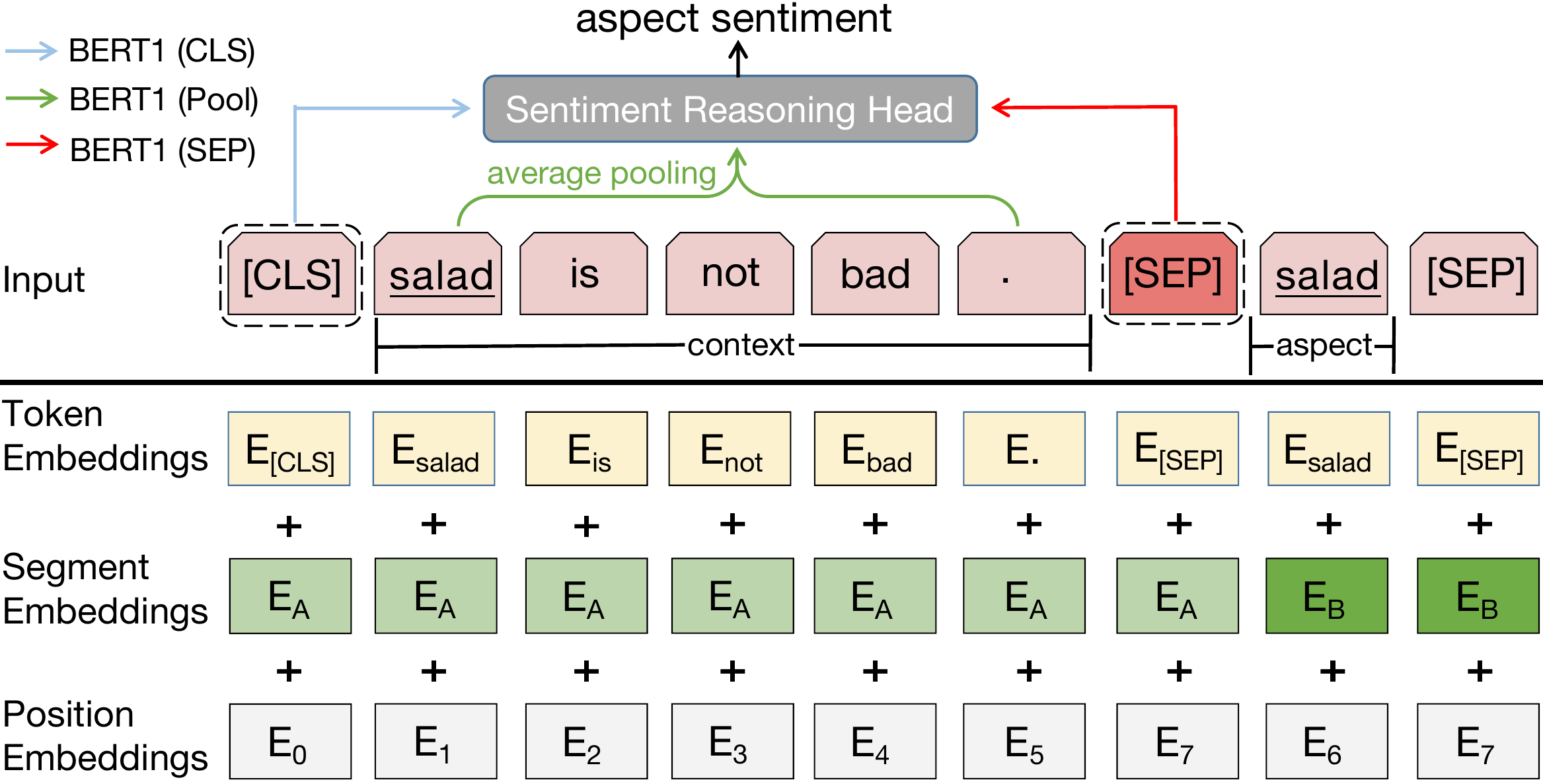}
 \caption{Illustration of BERT1 (CLS), BERT1 (Pool) and BERT1 (SEP). }
 \label{fig:bert1}
\end{figure}
Before we demonstrate our proposed AABERTs, we firstly introduce input and output formats of BERT0 and BERT1, as illustrated in Fig. \ref{fig:bert0} and Fig. \ref{fig:bert1}.
BERT (CLS) (BERT (SEP)) denotes the variant using the hidden state of \texttt{[CLS]} (\texttt{[SEP]}) token for classification.
BERT (Pool) uses the average pooling of all generated hidden states of context tokens for classification.
In the pre-training process of BERT, the hidden state of \texttt{[CLS]} token is used for prediction \cite{bert}.
To keep the fine-tuning consistent with the pre-training, previous works adopted BERT (CLS) rather than BERT (Pool) and BERT (SEP) as a strong baseline.
In this work, we not only study BERT (CLS) but also BERT (Pool) and BERT (SEP) for ABSA task.
As for the Sentiment Reasoning Head, we simply use a fully connected layer, whose output is the sentiment class probability distribution formed in a 3-d vector.

When designing AABERTs, the original language modeling capacity of pre-trained BERT should be preserved, so we do not modify its internal network architecture.
However, to achieve the aspect-awareness in BERT, we have to introduce the semantics of the given aspect to the intra-sentence dependency modeling process of BERT.
In this work, our intuition is to provide BERT with the signal of the given aspect without thoroughly separating the context and aspect as two individual sentences.
To achieve this, we try to break the isolation of the context ($s_1$) and the concatenated aspect ($s_2$), providing a more appropriate signal of the given aspect for BERT.
We modify the input format of BERT, more specifically, the setting of segment embedding and \texttt{[SEP]} token.
Note that although there are many works aiming to improve BERT, most of them focus on designing different pre-training tasks \cite{roberta,ernie,spanbert} and position embeddings\cite{posembed}, while the segment embedding and \texttt{[SEP]} token are yet to be studied.
Next we introduce the proposed three aspect-aware BERTs whose differences only lie in the input formats.
\subsubsection{AABERT1}
\begin{figure}[ht]
 \centering
 \includegraphics[width = \textwidth]{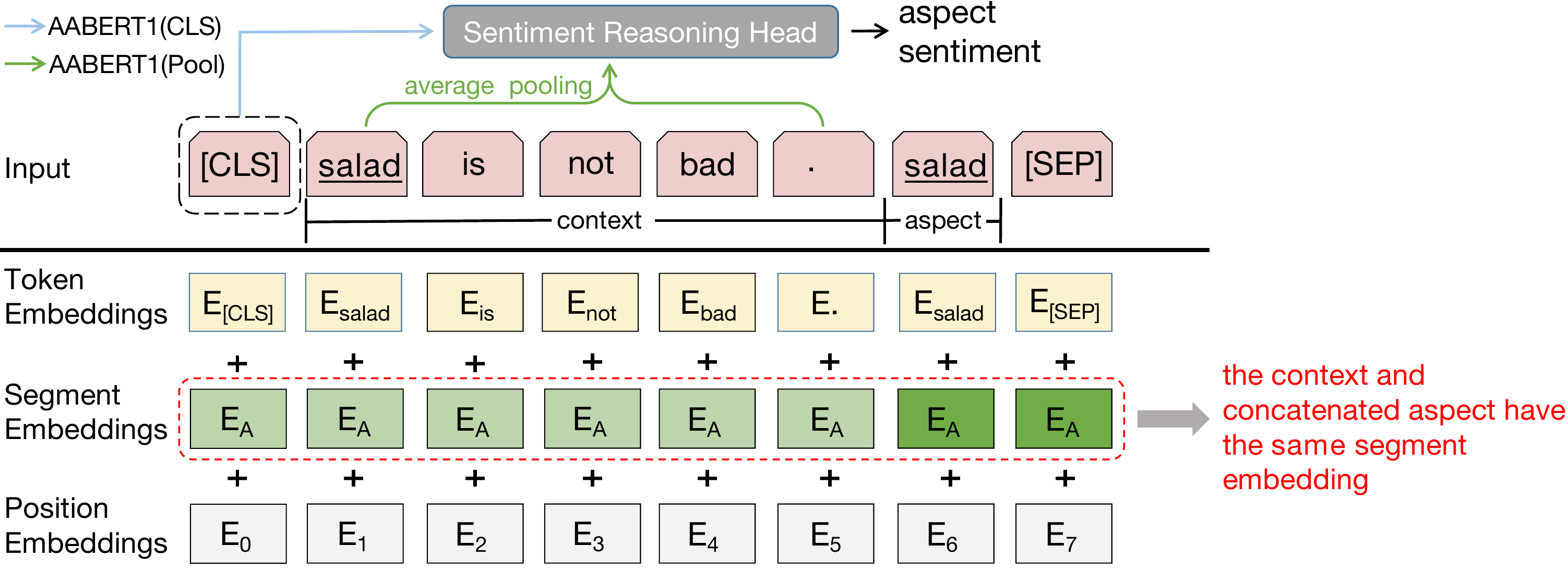}
 \caption{Illustration  of AABERT1 (CLS) and AABERT1 (Pool).}
 \label{fig:aabert1}
\end{figure}
The input and output format of AABERT1 is illustrated in Fig. \ref{fig:aabert1}.
In AABERT1, the context word sequence is concatenated with the word sequence of the given aspect.
They are not separated by \texttt{[SEP]} token and the segment embeddings of their tokens are identical.
As a result, the input tokens of context and aspect are not separated in the embedding space.
The intuition of AABERT1 is to provide the signal of the given aspect without separating the context and aspect.
There are two clues of aspect-awareness for intra-sentence context modeling.
First, considering the review is of one sentence, the punctuation mark (always `.') at the end of every review is a weak separator to hint that the tokens between it and the end mark (\texttt{[SEP]}) are the given aspect.
Second, the co-occurrence of the aspect included in the context ($s_1$) and the aspect in position $s_2$.

\subsubsection{AABERT2}
\begin{figure}[ht]
 \centering
 \includegraphics[width = \textwidth]{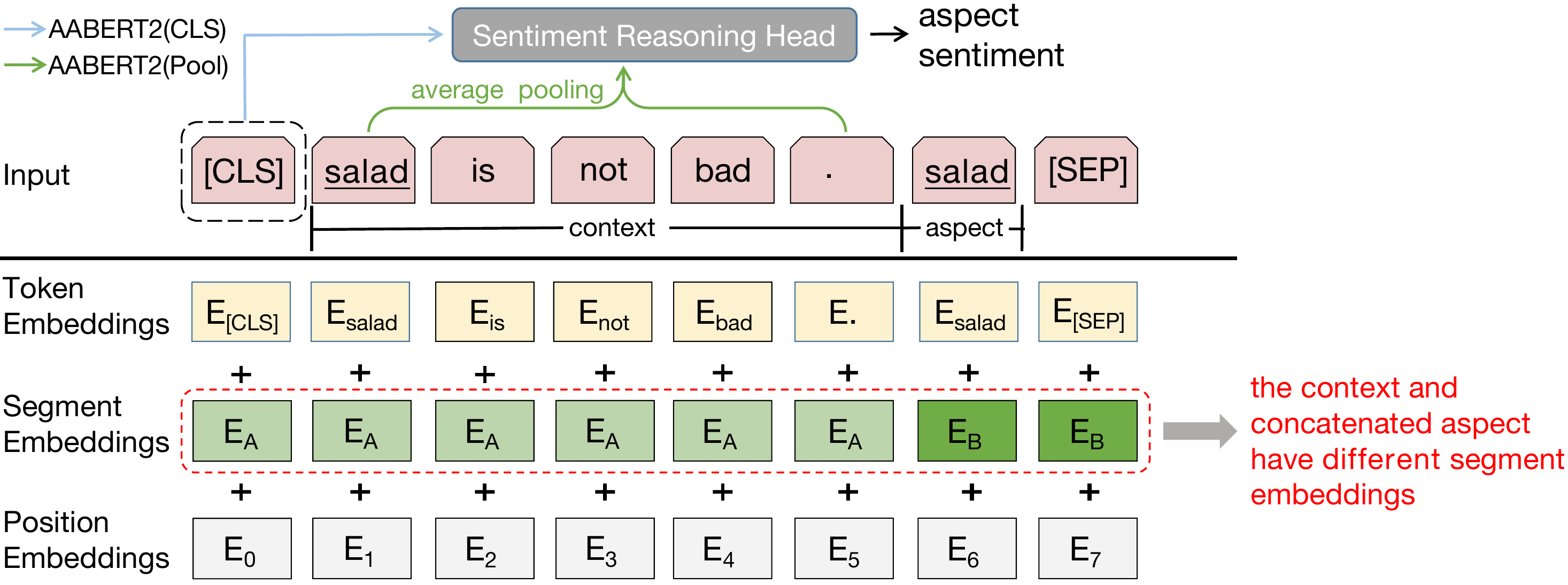}
 \caption{Illustration  of AABERT2 (CLS) and AABERT2 (Pool). }
 \label{fig:aabert2}
\end{figure}
The input and output format of AABERT2 is illustrated in Fig. \ref{fig:aabert2}.
The difference between AABERT2 and AABERT1 lies in the segment embedding.
In AABERT2, the segment embeddings of aspect tokens are set different from the ones of context tokens.
In the input embedding space, a signal is introduced to separate context and aspect.
Therefore, there are three clues indicating the specific aspect: (1) the final punctuation mark of the review context; (2) the co-occurrence of the concatenated aspect and the one include in the context; (3) the different semantic signals of the segment embeddings.
These three clues indicate the specific aspect of the current sample.
And more importantly, the concatenated aspect is not fully regarded as a separated sentence from the context.
Accordingly, the intra-sentence dependencies between the aspect and its related context words can be captured, then the generated hidden states can contain more useful aspect-related information.
\subsubsection{AABERT3}
\begin{figure}[ht]
 \centering
 \includegraphics[width = 0.7\textwidth]{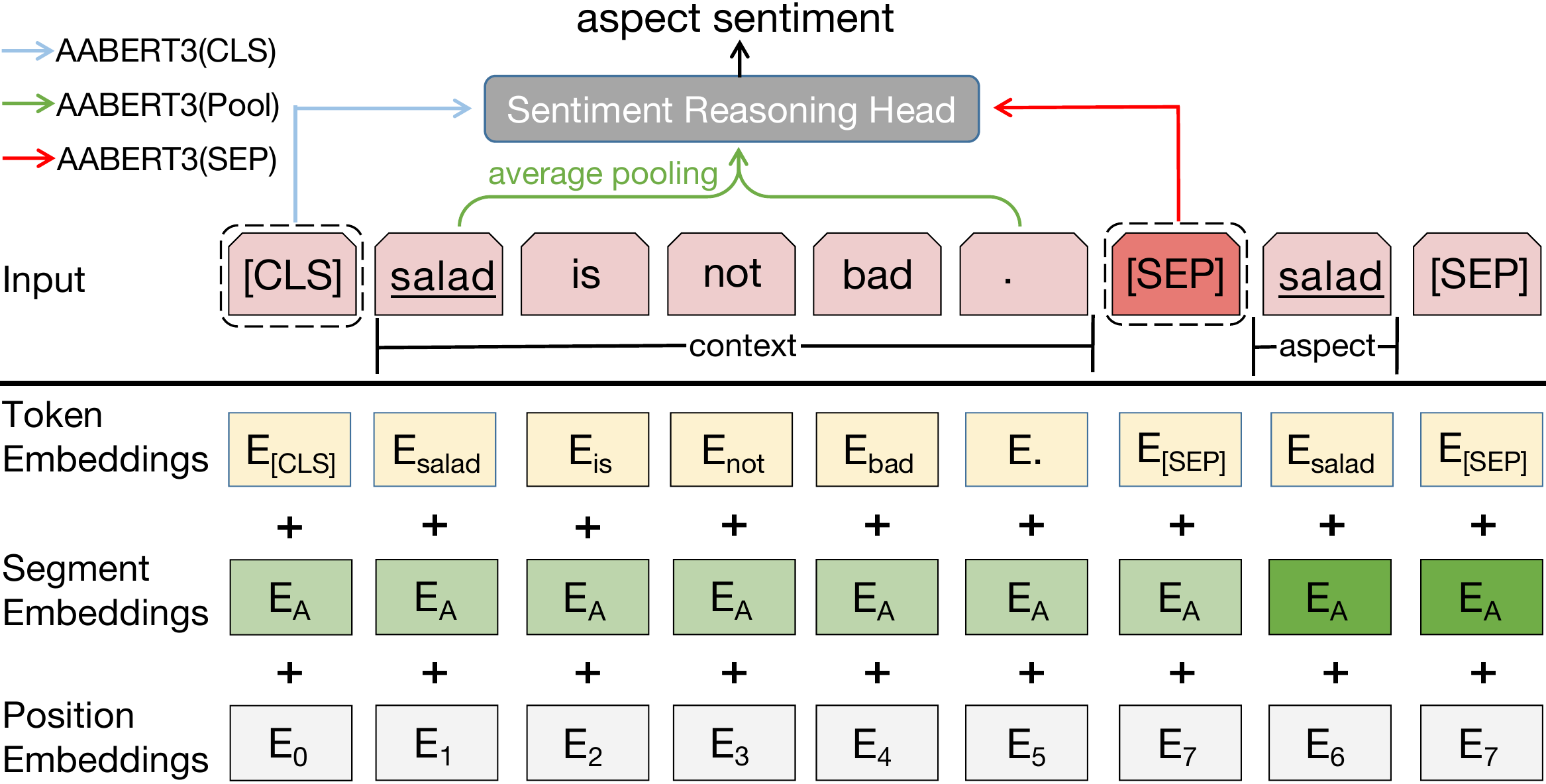}
 \caption{Illustration of AABERT3 (CLS), AABERT3 (Pool) and AABERT3 (SEP). }
 \label{fig:aabert3}
\end{figure}
The input and output format of AABERT3 is illustrated in Fig. \ref{fig:aabert3}.
AABERT3 adopts the explicit separator token \texttt{[SEP]} to separate the context and aspect in the input concatenated sequence.
However, AABERT3 does not set different segment embeddings for context tokens and aspect tokens to separate them in latent space.
Hence, there are two clues of aspect-awareness for intra-sentence context modeling: one is the \texttt{[SEP]} token and another is the co-occurrence of the aspect included in the context and the concatenated aspect in position $s_2$.
Similar to AABERT1 and AABERT2, the concatenated aspect is not fully regarded as a separated sentence from the context.
Thus the intra-sentence dependencies between the aspect and its related context words can be captured,

%% file: experiment.tex
\section{Experiments} \label{experiment}
\subsection{Task Definition}
We conduct experiments on the two cases of aspect-based sentiment analysis: aspect term sentiment analysis (ATSA) and aspect category sentiment analysis (ACSA).
The former infers sentiment polarities of the given aspect term explicitly included in the context word sequence.
The latter infers sentiment polarities of generic aspects such as \textit{service} or \textit{food} which may not be explicitly found in the context word sequence, and these generic aspects belong to a predefined set.
In this paper, both of ATSA and ACSA are classification tasks, concretely, they are respectively the subtask 2 (ST2) and subtask 4 (ST4) in SemEval-2014 Task 4 \cite{Semeval2014}.
\subsection{Settings}
\noindent \textbf{Datasets} \\We evaluate the performances of our models on SemEval 2014 \cite{Semeval2014} task 4 datasets which consist of \textit{laptop} and \textit{restaurant} reviews.
These two datasets are widely used benchmarks in many previous works \cite{ATAE,Tencent,tsmn,RGAT}, and consistent with them we remove the reviews having no aspect or the aspects with sentiment polarity of ``conflict''. Finally, the datasets consist of reviews with at least one aspect labeled with sentiment polarities of \textit{positive, neutral} and \textit{negative}.
For ATSA(ST2), we adopt \textit{Laptop} and \textit{Restaurant} datasets; for ACSA(ST4), only the \textit{Restaurant} dataset is available.
Full statistics of the datasets are given in Table 1.
\begin{table}[!hbtp]
\centering
\fontsize{10}{13}\selectfont
\linespread{0}
\caption{Statistic of all datasets}
\begin{tabular}{|l|l|r|r|r|}
\hline
Task                  & Dataset          & Pos  & Neg & Neu \\ \hline
\multirow{4}{*}{ATSA} & Restaurant Train & 2164 & 807 & 637 \\ \cline{2-5}
                      & Restaurant Test  & 728  & 196 & 196 \\ \cline{2-5}
                      & Laptop Train     & 994  & 870 & 464 \\ \cline{2-5}
                      & Laptop Test      & 341  & 128 & 169 \\ \hline{}
\multirow{2}{*}{ACSA} & Restaurant Train & 2179 & 839 & 500 \\ \cline{2-5}
                      & Restaurant Test  & 657  & 222 & 94  \\ \hline
\end{tabular}
\label{dataset}
\end{table}

\noindent \textbf{Evaluation}\\
We adopt both Macro-F1 and Accuracy (Acc) to evaluate the models' performances.
Generally, higher Acc can verify the effectiveness of the system while it biases towards the majority class, and Macro-F1 provides more indicative information about the average performance of all classes.

Following previous works \cite{asgcn,DGEDT,bigcn}, we train the models several times and report the average of best results of each run.
Concretely, we train the models 10 times with individual random seeds and the epoch number is 30.
\subsection{Base Models and Compared Models}
Firstly, we compare our AALSTM with LSTM as single models.
LSTM, LSTM (Pool) and Bi-LSTM denote that the last hidden state of LSTM, the pooling of all hidden states of LSTM and the pooling of all hidden states of Bi-LSTM are taken as the final representation for classification, respectively.
And the same notation for AALSTM, as shown in Table \ref{lstm-results}.
The output of Bi-AALSTM is the series of concatenated hidden states of two AALSTMs of different directions.

In Section \ref{sec:introduction}, we divide existing LSTM-based ABSA models into four categories according to their context modeling process.
 In order to verify the superiority of the proposed AALSTM and AAGCN compared to vanilla LSTM and GCN, we choose some representative models as backbones and replace their original vanilla LSTM and GCN with our proposed AALSTM and AAGCN.
  We select one representative model from each of these categories for experiments. Accordingly, ATAE-LSTM \cite{ATAE}, IAN \cite{IAN}, RAM \cite{Tencent}, and ASGCN \cite{asgcn} are chosen as the representatives of the four categories because their architectures are novel and they are widely taken as compared models in previous works. We introduce the four LSTM-based backbones as follows:
 \begin{itemize}
\item \textbf{Attention-based LSTM with Aspect Embedding (ATAE-LSTM).} It concatenates the aspect embedding to the word embeddings of context words as the initial context word representation input into the LSTM layer. In the attention mechanism, the aspect embedding is utilized to produce the attention vector. We refer to this model as ATAE for short.
\item \textbf{Interactive Attention Networks (IAN).} It models context and target separately. And in the interactive attention mechanism, for ATSA task, the context and target leverage the average of each other's hidden states to produce their attention vectors. The representations of the context and aspect are concatenated as the final aspect-based sentiment representations. For the ACSA task, the target modeling module is not available because the aspect category is not a word sequence as the target. So we use the aspect embedding to produce the attention weights of context words.
\item \textbf{Recurrent Attention Network on Memory (RAM).} Aiming at ATSA task, it utilizes aspect-relative location to assign weights to original hidden states then produces the attention vector in the recurrent attention mechanism consisting of GRU cells. It can not be applied to the ACSA task for that the aspect category has no location information because it is not included in the word sequence of context.
\item \textbf{Aspect-specific Graph Convolution Networks.} Aiming at ATSA task, it employs a GCN to encode the local connections of the syntax graph of context, and use an aspect-aware attention to extract the aspect-related semantics for classification. This model cannot be applied to the ACSA task either, for the same reason as RAM.
\end{itemize}

Except for evaluating BERT0, BERT1, and our three AABERTs as single models, we also compare them as context encoders in X+BERT experiments by replacing the original BERT encoder with AABERTs.
We choose R-GAT+BERT 
\cite{RGAT}, which is a recently proposed BERT-based model for ABSA, as our backbone for X+BERT experiments.
\begin{itemize}
\item \textbf{Relational Graph Attention Network with BERT Encoder.}
Augmenting graph attention network \cite{gat} with relation embedding, \cite{RGAT} propose relational graph attention network (R-GAT) which can capture the relations between the aspect and each context word via operating on the star-shaped aspect-oriented dependency graph.
R-GAT+BERT uses BERT1 to model context and aspect together in the sentence-pair manner and takes the output hidden state of \texttt{[CLS]} token as the aspect representation.
\end{itemize}

Some state-of-the-art (SOTA) models are chosen for comparison: 
\begin{itemize}
\item \textbf{ATSA (ST2):} TNet \cite{Lixin}; TD-GAT \cite{graphatt}; BiGCN \cite{bigcn}; R-GAT \cite{RGAT}; TDGEDT-BERT \cite{DGEDT}; AEN-BERT \cite{aen-bert}; KGCapsAN-BERT \cite{kgcap}; ASGCN + BERT \cite{asgcn}; A-KVMN+BERT \cite{kvmn-eacl}; BERT+T-GCN \cite{tgcn}.
\item \textbf{ACSA (ST4):} AS-Capsule \cite{absacap}, BERT-pair-QA-B \cite{bert-qa}.
\end{itemize}

For fair comparison, we leave out the baselines that utilizes external resources \cite{access2019,bert_post,chinese-oriented}.
\subsection{Implementation Details}
 For LSTM based models, we initialize all word embeddings by Glove vectors \cite{Glove} and the out-of-vocabulary words' embeddings are sampled from the uniform distribution $U(-0.1,0.1)$. 
 All embedding dimensions are set to 300.
 For BERT-based models, we adopt BERT-base uncased English version \cite{bert}.
 Initial values of all weight matrices are sampled from the uniform distribution $U(-0.1,0.1)$ and initial values of all biases are zeros.
The batch size is set as 16. We minimize the cross-entropy loss to train our models using the Adam optimizer \cite{Adam} with the learning rate set as 0.001 for LSTM-based models and $10^{-5}$ for BERT-based models. To avoid overfitting, for LSTM based models, we adopt the dropout strategy on the input context wording embedding layer and the final aspect-based sentiment representation with $p=0.5$; for BERT-based models, we adopt the dropout strategy on the BERT output hidden states with $p = 0.3$. 
Besides, weight decay strategy and \textit{L2}-regularization are respectively adopted for BERT and LSTM based models.

As for the aspect vector ($A$) input into AALSTM or AAGCN, we set it as follows:
 \begin{itemize}
\item \textbf{ATSA (ST2):} \\
For AALSTM, generally we use the average of word embeddings of the target words as $A$, while for IAN, we adopt the average of the hidden states of the aspect words as $A$.
For AAGCN, the average pooling of the hidden states of the aspect is taken as $A$.
\item \textbf{ACSA (ST4):}\\
 For all models, we initialize all aspect embeddings by sampling from the uniform distribution $U(-0.1,0.1)$.
\end{itemize}
For BERT-based models, the aspect category is regarded as a phrase just like the aspect term which is concatenated to the end of context in the input sequence.

%% file: result.tex
\section{Main Results} \label{mainresult}
\begin{table*}[th]
\centering
\fontsize{8}{10}\selectfont
\setlength{\belowcaptionskip}{2pt}
\caption{Evaluation results of LSTM-based models. Best scores are in \textbf{bold}. (AA) denotes that the vanilla LSTM in the backbone  is replaced with our AALSTM; (Bi) denotes that the LSTM used in the backbone is Bi-LSTM; (Bi-AA) denotes that the LSTM used in the backbone is replaced with our Bi-AALSTM. $^\dag$  indicates that the aspect-aware variant significantly outperforms the corresponding backbone, with $p < 0.05$ under t-test. }
\setlength{\tabcolsep}{0.7mm}{
\begin{tabular}{c|c|c|cc|cc|cc}\toprule
\multicolumn{2}{c|}{\multirow{2}{*}{Models}} & \multirow{2}{*}{Aspect-Aware} & \multicolumn{2}{c|}{ATSA(ST2)-Res14} & \multicolumn{2}{c|}{ATSA(ST2)-Lap14} & \multicolumn{2}{c}{ACSA(ST4)-Res14}  \\ \cline{4-9}
\multicolumn{2}{c|}{} &  & Macro-F1            & Accuracy            & Macro-F1         & Accuracy            & Macro-F1         & Accuracy   \\ \midrule
\multirow{6}{*}{\rotatebox{90}{Single LSTM}}
&LSTM       & - & 61.17  &75.93  & 60.82 & 68.57  &71.66  &82.15 \\
&LSTM (Pool)  & - & 61.22 & 76.01 & 60.75 & 68.71  & 71.84 & 82.26 \\
&Bi-LSTM (Pool)  & -  & 62.45 & 76.61 & 61.33 & 68.94 & 72.07 &82.69 \\
&AALSTM$^\dag$      & AA & 65.79  &78.09  & 63.58 & 70.41  &75.23  &84.56 \\
&AALSTM (Pool)$^\dag$   &AA  &66.21 & 78.22 & 63.76 &71.54 & 75.55 &84.80 \\
&Bi-AALSTM (Pool)$^\dag$   &Bi-AA &68.91 & 79.42& 64.25&72.21 &76.32 & 85.23 \\
 \midrule
\multirow{18}{*}{\rotatebox{90}{X + LSTM}}
&ATAE      &- & 63.12  &76.83 & 60.56 & 66.96  &70.61  &82.17 \\
&ATAE (AA) &AA & 63.11  &76.24  & 58.17 & 65.39  &71.04  &82.08 \\
&ATAE (Bi)  &- &64.99 & 77.41& 62.81&67.40 &71.84 &81.71 \\
&ATAE (Bi-AA) & Bi-AA &64.27 & 76.43&62.64 &68.31 &71.38&82.11 \\
\cline{2-9} 
&IAN   & -   & 65.70  &78.39  & 64.32 & 71.32  &73.00  &83.66 \\
&IAN (AA)$^\dag$ &AA     & 67.66  &78.69  & 67.13 & 72.54  &75.55  &84.86\\
&IAN (Bi) & - &68.44 & 79.02& 65.74 &72.10 &73.43 &83.86 \\
&IAN (Bi-AA)$^\dag$   &Bi-AA  &69.35 & 79.86& 68.11 & 73.25&\textbf{76.45} &\textbf{85.41} \\
\cline{2-9} 
&RAM   & -   & 68.47  & 78.21 &67.89 &72.41  &-  &- \\
&RAM (AA)$^\dag$   &AA   & 69.54 & 79.16  & 70.41  &74.46    &-  &- \\
&RAM (Bi) & - & 67.78  &79.46  & 68.58 & 73.35  &-  &- \\
&RAM (Bi-AA)$^\dag$  &Bi-AA &71.27 & 80.21&71.15 &75.12 &- &- \\
&RAM (Bi-AA) + AAGCN$^\dag$   &Bi-AA + AAGCN &\textbf{75.17} &\textbf{82.53} &\textbf{73.77} &\textbf{77.27} &- &- \\
\cline{2-9}
&ASGCN (Bi)  & -   & 72.02  &80.77  & 71.05 & 75.55  &-  &- \\
&ASGCN (Bi-AA)$^\dag$    &Bi-AA    & 73.51  &81.78  & 72.96 & 76.33  &-  &- \\
&ASGCN (Bi, AAGCN)$^\dag$    &AAGCN    & 73.74& 81.96&  72.34 & 76.96  &-  &- \\
&ASGCN (Bi-AA, AAGCN)$^\dag$   &Bi-AA + AAGCN   & 74.14  &82.18 & 73.44& 77.22& -&- \\ \midrule \midrule
\multirow{5}{*}{\rotatebox{90}{SOTA}}
&TNet &- &71.03&80.42&70.14&74.61&-&- \\
&TD-GAT &- &71.72&81.32&70.74&75.63&-&-\\
&BiGCN  & -  &  73.48   & 81.97    &71.84     &74.59    & -    & - \\
&R-GAT &- & 73.52  &  81.55 & 72.11 & 75.52 & -&- \\
&AS-Capsule & -&-&-&-&- &73.53 &85.0\\
\bottomrule
\end{tabular}}
\label{lstm-results}
\end{table*}

\subsection{LSTM}
Experimental results of LSTM-based models are illustrated in Table \ref{lstm-results}.
First of all, we can find that single AALSTM significantly outperforms vanilla LSTM and ATAE, and even close to IAN and RAM.
On \textit{ATSA(ST2)-Res14} dataset, AALSTM overpass LSTM by 4.6\% on F1 score.
It is worth mentioning that ATAE, IAN, and RAM all adopt the attention mechanism to extract aspect-related information, while AALSTM and LSTM only model the context and take the last hidden state for prediction. 
The satisfying performances of single AALSTM prove that its generated last hidden state $H$ contains better and more sentiment indicative information of the given aspect than the last hidden state generated by vanilla LSTM.
Generally, the location of the tokens mentioning the given aspect is not at the end of the review context.
So it is proved that after training, AALSTM is able to transmit the aspect-related semantics in different context words to the last hidden state along time steps.
And we can observe that generally the pooling variants of LSTM and AALSTM perform better than the ones adopting the last hidden state for classification.
We speculate that there is some important information which is not contained in the last hidden state, and the pooling variant can aggregate the important in all hidden states into the final representation, which leads to better performance.
Besides, we can find that Bi-AALSTM obtains significant improvement via only concatenating two unidirectional AALSTM.
This is because Bi-AALSTM can aggregate the aspect-specific semantics from both directions, and the final representation is more sufficient.

To further verify the superiority of AALSTM over vanilla LSTM, we replace the original LSTM in the four backbones (ATAE, IAN, RAM and ASGCN) with our proposed AALSTM. 
In the implementation of our experiments, the only difference between original models and their AA or Bi-AA variants is the replacement of vanilla LSTM or Bi-LSTM. 
So the performance improvement can directly demonstrate the effectiveness of our proposed AALSTM and Bi-AALSTM.
We can observe that (Bi-)AALSTM can significantly improve the performances of IAN, RAM and ASGCN, especially on Macro-F1.
On \textit{ATSA(ST2)-Lap14} dataset, IAN (AA) has 2.8\% and 1.2\% improvement on F1 and Acc, respectively.
On ACSA task, IAN (AA) gets an improvement of 2.6\% on F1 compared with IAN, and IAN (Bi-AA) outperform IAN (Bi) on ACSA(ST4) by 3\% in terms of F1.
IAN (AA) and IAN (Bi-AA) outperforms AS-Capsule by 3\% on F1.
Compared with AS-Capsule,  IAN (AA) and IAN (Bi-AA) respectively obtain 2\% and 3\% improvement in terms of F1.

\begin{figure*}[t]
 \centering{}
 \includegraphics[width = 1\textwidth]{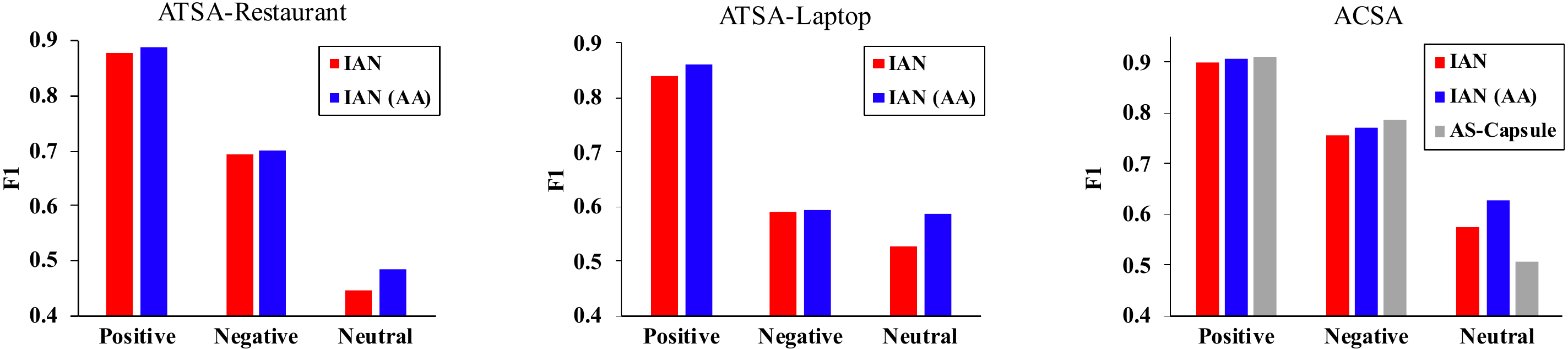}
 \caption{Performance comparison on different sentiment polarities.}
 \label{chart-pol}
\end{figure*}

To analyze the significant improvements on F1, we dissect the performances of IAN and IAN (AA) on the three sentiment classes, and AS-Capsule is also compared on ACSA task.
The comparisons are shown in Fig. \ref{chart-pol}.
We can observe that AALSTM can improve IAN’s performance on all sentiment classes. Especially, AALSTM can significantly improve the performance on \textit{Neutral} class.
And IAN (AA) overpasses AS-Capsule on \textit{Neutral} class by a large margin.
As shown in previous works \cite{ATAE,Fusion,absacap}, the sentiment prediction of \textit{Neutral} class is much harder than the other two sentiment classes.
This is caused by two main issues.
First, as shown in Table \ref{dataset}, the \textit{Neutral} class has much fewer samples, compared with the other two sentiment classes. 
This causes neural ABSA models do not have enough chances to learn how to effectively identify and extract the features that are important for correctly inferring the \textit{Neutral} sentiment.
Second, the key information about the given aspect with \textit{Neutral} sentiment may be discarded in the context modeling process of vanilla LSTM, especially on the multi-aspect situation where there are other aspects with \textit{Positive} or \textit{Negative} sentiment. 
Compared with the tokens expressing \textit{Positive} or \textit{Negative} sentiments on other aspects in the same review, the tokens modifying \textit{Neutral} aspect is prone to be neglected by vanilla LSTM cells to some extent.
This is because the semantic information of the tokens modifying \textit{Neutral} aspect seems too general against the overall semantic background of the context. 
The comparisons in Fig \ref{chart-pol} proves that by taking the semantics of the given aspect into consideration when modeling context, AALSTM can effectively alleviate above two issues.

However, we can find that AALSTM brings diminishing performances for ATAE model.
ATAE represents a category of models that utilize aspect embedding concatenation (AEC) for the joint modeling of aspect and context.
We suspect that the aspect information fused in the context word representations input into AALSTM may adversely affect the aspect-aware mechanism's ability of identifying the sentiment indicative information of the given aspect and the noise information with respect to the given aspect (this conjecture is verified in Sect. \ref{diss-ae}).

To investigate the efficacy of our proposed AAGCN, we replace the vanilla GCN in ASGCN with it.
We can observe that ASGCN (Bi, AAGCN) significantly outperforms ASGCN (Bi).
This proves the superiority of AAGCN over vanilla GCN.
The aspect-aware graph convolution process of AAGCN can aggregate more aspect-related semantics to current node from its neighbor nodes, and discard the useless information simultaneously.
Therefore, AAGCN can generate better node representations, which improve performances.
With the adoption of both Bi-AALSTM and AAGCN, ASGCN (Bi-AA, AAGCN) obtains even better performances, outperforming up-to-date state-of-the-art models by large margins.
This proves that AALSTM can work well with AAGCN, generating much better hidden states which contain not only  aspect-related semantics but also syntactic information.
Additionally, we add AAGCN to RAM (Bi-AA) and we can find the performances are significantly improved, even overpassing ASGCN (Bi-AA, AAGCN).
This is because the addition of AAGCN not only introduces syntactic information but also further regulates the information flow regarding the specific aspect.
\subsection{BERT}
\begin{table}[ht]
\centering
\caption{Illustration of input formats of different BERT versions. 'SEP Token' column indicates whether using the [SEP] token. 'Segment Embedding' column indicates whether the segment embeddings of context words and the concatenated aspect words are different. }
\fontsize{10}{14}\selectfont
\begin{tabular}{|l|c|c|c|}
\hline
Models  & SEP Token & Segment Embedding & Input Format \\ \hline
BERT0   &   -        &        -           &    [CLS]+ Context  +[SEP]       \\ \hline
BERT1   &   $\surd$      &      $\surd$        &[CLS]+  Context + [SEP] + Aspect + [SEP]             \\ \hline
AABERT1 &     -      &         -          &   [CLS]+  Context + Aspect + [SEP]            \\ \hline
AABERT2 &      -     &      $\surd$    &   [CLS]+  Context + Aspect + [SEP]           \\ \hline
AABERT3 &    $\surd$   &      -     &    [CLS]+  Context + [SEP] + Aspect + [SEP]          \\ \hline
\end{tabular}
\label{tabel: format}
\end{table}
\begin{table*}[th]
\centering
\caption{Evaluation results of BERT-based models. Best scores are in \textbf{bold}. $^\dag$ indicates that the score significantly overpass the baselines', with $p < 0.05$ under t-test.}
\fontsize{9}{12}\selectfont
\setlength{\tabcolsep}{0.7mm}{
\begin{tabular}{c|l|cc|cc|cc}\toprule
\multicolumn{2}{c|}{\multirow{2}{*}{Models}} & \multicolumn{2}{c|}{ATSA(ST2)-Res14} & \multicolumn{2}{c|}{ATSA(ST2)-Lap14} & \multicolumn{2}{c}{ACSA(ST4)-Res14}  \\ \cline{3-8} \addlinespace[0.5ex]
\multicolumn{2}{c|}{}  & Macro-F1            & Accuracy            & Macro-F1         & Accuracy            & Macro-F1         & Accuracy   \\ \midrule
\multirow{12}{*}{\rotatebox{90}{Single BERT}}
&BERT0 (CLS)       & 71.26  &81.93  & 73.61 & 78.12  &81.03  &88.90 \\
&BERT0 (Pool)     & 72.85  &82.46  & 73.40 & 77.99  &81.93  &89.42\\
& BERT1 (CLS)     & 79.68  &85.96  & 74.70 & 78.79  &84.20  &90.61    \\
& BERT1 (Pool)    & 80.07  &86.16  & 76.27 & 80.06  &84.65  &90.69    \\
& BERT1 (SEP)     & 79.90  &86.09  & 75.84 & 79.48  &85.16  &91.20\\
& AABERT1 (CLS)     & 80.17  &86.27  & 75.71 & 79.91  &85.18  &90.96\\
& AABERT1 (Pool)    & 79.75  &86.06  & 76.19 & 80.03  &83.35  &89.89\\
& AABERT2 (CLS)      & 80.08  &86.19  & 75.88 & 79.61  &84.42  &90.79\\
& AABERT2 (Pool)    & 79.98  &86.04  & 76.61 & 79.98  &83.79  &90.28\\
& AABERT3 (CLS)      & 79.90  &85.97  & 75.41 & 79.31  &84.65  &90.76\\
& AABERT3 (Pool)    & 79.76  &86.09  & 76.27 & 79.67  &84.67  &90.65\\
& AABERT3 (SEP)     & 79.42  &85.86  & 74.76 & 78.68  &84.85  &90.64\\
\midrule
\multirow{10}{*}{\rotatebox{90}{X + BERT}}
&R-GAT + BERT0    & 73.51  &82.48  & 73.75 & 78.10  &-  &-\\
&R-GAT + BERT1    & 80.23  &86.22  & 75.71 & 79.45  &-  &-\\
&R-GAT + AABERT1    & 80.12  &86.19  & 76.07 & 79.91  &-  &-\\
&R-GAT + AABERT2    & 79.82  &86.07  & 76.00 & 79.83  &-  &-\\
&R-GAT + AABERT3    & 79.55  &85.99  & 75.78 & 79.56 &-  &- \\ \cline{2-8}
&R-GAT2 + BERT0   & 79.60  &86.03  & 75.76 & 79.37  &-  &-\\
&R-GAT2 + BERT1   & 80.31  &86.44  & 75.84 & 79.37  &-  &-\\
&R-GAT2 + AABERT1   & 80.21 &86.38  & 76.22 & 79.84  &- &-\\
&R-GAT2 + AABERT2   & 80.07  &86.46 & 76.21 & 79.73  &- &-\\
&R-GAT2 + AABERT3   & 80.47  &86.59  & 76.32 & 79.76  &-&-\\ \cline{2-8}
&AAGCN + AABERT3 &-&-&-&-& \textbf{86.74}$^\dag$&\textbf{92.82}$^\dag$\\ \cline{2-8}
&ASGCN (AAGCN) + AABERT3 &\textbf{80.84}$^\dag$ &86.64 &77.20 &80.56 &- &- \\ \cline{2-8}
&RAM + AAGCN + AABERT3 &80.69 &\textbf{86.79}$^\dag$ & \textbf{78.01}$^\dag$&\textbf{81.20}$^\dag$ &- &- \\ \hline \hline
\multirow{7}{*}{\rotatebox{90}{SOTA}}
&AEN-BERT  &  73.76   &83.12  & 76.31 &   79.93  & -    &  -   \\
&DGEDT-BERT   &  80.0    &86.3    &  75.6    & 79.8   & -  &-    \\
&KGCapsAN-BERT   & 79.00    &85.36     &   76.61   &  79.47  & - & - \\
&ASGCN+BERT     &  79.32   &85.87      & 74.35  & 78.92  & - & -        \\
&A-KVMN+BERT   &  77.94   &85.98    &  76.14   &  79.78  & -  &-    \\
&BERT+T-GCN    &  79.40  & 85.95   &   76.95   &  80.56   &- & - \\
&BERT-pair-QA-B &-&-&-&- &- & 89.9\\
\bottomrule
\end{tabular}}
\label{bert-results}
\end{table*}
We list the input formats of BERT0, BERT1 and our three AABERTs in Table \ref{tabel: format}  to help distinguish them.
Experimental results of BERT-based models are illustrated in Table \ref{bert-results}.
We can observe that BERT0 (CLS) and BERT0 (Pool) obtain the worst results. This is because there is no aspect semantics introduced.
And among other single BERT models, BERT1 (CLS) achieves the worst results on ATSA task, although it is widely used for sentence classification.
Our proposed AABERTs show significant superiority to BERT1 on ATSA task.
Remarkably, AABERT2 (Pool) even overpass some state-of-the-art models in terms of F1 on \textit{ATSA(ST2)-Lap14} dataset.
This proves its strong capability of extracting aspect-related information from context words, which is because AABERT1 can effectively model the \textbf{intra-sentence dependencies} between the given aspect and context words.
On ACSA task, we can find that BERT1 (SEP) slightly outperforming our proposed AABERTs.
This is because aspect categories are not included in the context word sequence.
In this case, AABERTs' advantage of intra-sentence dependencies modeling is hindered because the aspect is not a word sequence included in the sentence.

To evaluate the effectiveness of AABERTs as context encoders, we conduct experiments on recent proposed R-GAT+BERT by replacing the original BERT with AABERTs.
Note that R-GAT+BERT is only available for ATSA task and in its original architecture, the BERT is BERT1 and the hidden state of \texttt{[CLS]} token is taken as the aspect representation.
We can find that generally, R-GAT+AABERTs outperform R-GAT+BERT1 and R-GAT+BERT0, while on \textit{ATSA-Res14} dataset R-GAT+AABERTs slightly underperform to R-GAT+BERT1.


As R-GAT+BERT takes the hidden state of \texttt{[CLS]} token as aspect representation, we suspect the generated hidden states of \texttt{[CLS]} tokens may lose some important aspect-related information when aggregating the (aspect-related) context information.
Since average pooling can retain the original information as much as possible, we
propose R-GAT2+BERT which takes the average pooling of all context hidden states as aspect representations.
The performances of R-GAT2+BERTs are demonstrated in Table \ref{bert-results}.
We can observe that the performances are improved via only changing the way of obtaining the aspect representation from CLS to Pool .
In addition, this operation costs no extra time and computation at implementation.

Another interesting phenomenon is that the collaboration of R-GAT and AABERTs (BERT0/BERT1) does not bring improvements for AABERTs (BERT1) on \textit{ATSA(ST2)-Lap14} dataset.
We suspect the reason is that the grammatical correctness of the reviews in \textit{ATSA(ST2)-Lap14} dataset is lower than the reviews in \textit{ATSA(ST2)-Res14} dataset, which was revealed in \cite{walk}.
And the R-GAT heavily relies on the precision of dependency graphs, which may be incorrectly parsed due to the ungrammatical sentences.
As a result, the performances of R-GAT+BERTs on \textit{ATSA(ST2)-Lap14} dataset are not promising enough.

To evaluate the efficacy of combining AAGCN and AABERT, we design three models: AAGCN + AABERT3, ASGCN (AAGCN) + AABERT3 and RAM + AAGCN+ AABERT3, in which the original LSTM encoder is replaced with our proposed AABERT3 encoder.
AAGCN + AABERT3 takes the average pooling of hidden states as the final representation for classification.
We can observe that it achieves new state-of-the-art performances on ACSA(ST4).
This proves that although AABERT have strong ability of context modeling and retain aspect-related semantics, AAGCN can further improve the performance by integrating syntactic information and further aggregating aspect-centric semantics.
And we can find that RAM + AAGCN + AABERT3 achieves new state-of-the-art performance on ATSA(ST2).
Although RAM is a model proposed in 2017, with the power of our proposed AAGCN and AABERT3, it beats the up-to-date models proposed in 2020 and 2021.

It is worth mentioning that all of the improvements of AABERTs are achieved by skillfully modifying the input format of BERT, which is pretty easy to implement in practice.
And our proposed AABERTs do not cause extra computation and time cost because they have the same architecture and parameters with vanilla BERT.
Duo to this, the improvements of AABERTs over vanilla BERT are not very significant.
However, AAGCN can work well with AABERTs to generate high-quality hidden states which can significantly improve performances and lead to new state-of-the-art results.

\subsection{Computation Efficiency of AALSTM- and AABERT-based Models}
From the results demonstrated above, we can observe than BERT has absolute superiority over LSTM: even the basic BERT1 significantly overperforms the best LSTM-based model RAM (Bi-AA) + AAGCN.
However, this does not mean that LSTM-based models should be abandoned.
In real-world scenarios, not only the model performance is important, but also computation efficiency is also crucial.
Table \ref{table: computation} shows the Computation Efficiency of our proposed RAM (Bi-AA) + AAGCN and RAM + AAGCN + AABERT3.
We can find that the AABERT-based model costs 3 times of training and inference time compared with the AALSTM-based model, and even requires 6 times of GPU memory, while it only surpasses the AALSTM-based model by about 5\% on performance.
Therefore, LSTM-based models is much more efficient on computation and costs much less resource.
In some cases, especially the ones demanding fast inference while without adequate computing resources, BERT-based models are impractical to use.
In contrast, LSTM-based models can effectively handle the task with much fewer parameters and faster inference speed.

\begin{table}[t]
\centering
\caption{Comparison of our proposed LSTM- and BERT- based models on different factors of Computation Efficiency.}
\fontsize{10}{14}\selectfont
\begin{tabular}{|l|l|l|l|l|}
\hline
Models                & \tabincell{c}{Training Time \\(per epoch)} & \tabincell{c}{ Inference Time \\(per sample)} & GPU Memory & Acc    \\ \hline
RAM (Bi-AA) + AAGCN   & 4.8s                      & 0.7ms                       & 1.2G       & 77.3\% \\ \hline
RAM + AAGCN + AABERT3 & 14.4s                     & 2ms                         & 7.2G       & 81.2\% \\ \hline
Improvement of BERT   & 300\%                     & 300\%                       & 600\%      & 5\%    \\ \hline
\end{tabular}
\label{table: computation}
\end{table}

%% file: discussion.tex
\section{Discussion} \label{discussion}
\subsection{Effect of Aspect Embedding Concatenation} \label{diss-ae}

In Section \ref{mainresult}, we observe that the ATAE (AA) performs worse than ATAE.
And we speculate the reason is the aspect embedding concatenation (AEC) adversely affects the aspect-aware gate mechanism of AALSTM, which is characterized by its ability of identifying the beneficial and adverse information with respect to given aspect at every time step.
From Eq. 1, 3, and 7 we can find that the aspect gate mechanism is controlled by $A$ and $h_{t-1}$ which are concatenated together.
The objective of the aspect-aware gate mechanism is to model the relation between the given aspect and the current semantic state then determine how much information from the given aspect should be integrated into the three gates of LSTM.
But because of AEC, much information of the given aspect is fused into the generated hidden states.
So $A$ and $h_{t-1}$ have much overlap with each other.
This greatly increases the difficulty of modeling their relation.
Moreover, from Eq. 2, 4, 8 we can find that the weighted $A$ is added to the internal value of the three gates of LSTM.
The object of this operation is to regulate the information flow of LSTM cells by adding some disturbance from $A$ to the original internal value calculated by $x_t$ and $h_{t-1}$.
Without AEC, there is little aspect information in $x_t$ and $h_{t-1}$, so the disturbance from $A$ can effectively adjust the three gates of LSTM with respect to the given aspect.
But if AEC is adopted, there will be much aspect information contained in $x_t$ and $h_{t-1}$.
As a result, the influence of the disturbance from $A$ is little, even invalid from the experimental results.
\begin{table*}[h]
\centering
\caption{Performance comparison of ATAE variants on three datasets. $^\dag$ indicates that ATAE (AA) w/o AEC significantly outperform ATAE and ATAE (AA), with $p < 0.05$ under t-test.}
\fontsize{8}{11}\selectfont
\setlength{\tabcolsep}{1.5mm}{
\begin{tabular}{l|cc|cc||cc}\toprule 
\multirow{2}{*}{Models}
           & \multicolumn{2}{c|}{ATSA(ST2)-Res14} & \multicolumn{2}{c||}{ATSA(ST2)-Lap14} & \multicolumn{2}{c}{ACSA(ST4)-Res14}  \\ \cline{2-7} \addlinespace[0.5ex]
  & Macro-F1            & Accuracy            & Macro-F1         & Accuracy            & Macro-F1         & Accuracy   \\ \midrule

ATAE       & 63.12  &76.83  & 60.56 & 66.96  &70.61 &82.17 \\
ATAE (AA)  & 63.11  &76.24  & 58.17& 65.39  &71.04  &82.08 \\
ATAE (AA) w/o AEC   & 65.08$^\dag$    &77.10$^\dag$    & 62.45$^\dag$   & 68.87$^\dag$    &72.61$^\dag$    &83.48$^\dag$  \\
\bottomrule
\end{tabular}}
\label{noAE}
\end{table*}

To confirm our conjecture, we conduct a set of ablation experiments based on ATAE model to study the effect of AEC.
Table \ref{noAE} demonstrates the experimental results.
We can find that if AEC is removed from ATAE (AA), the model's performance gets significant increase.
Just because of the adoption of AEC, the advantage of AALSTM is blocked out.
The evaluation results of this set of ablation experiments provide solid evidence that AEC has an adverse influence on AALSTM.
Based on the experimental results we can come to the conclusion that the \textit{purity} of input initial context word representations is critical for AALSTM to retain the important information and eliminate the adverse information according to the given aspect for ABSA task.
The \textit{purity} here means the input word embeddings to AALSTM should only contain the semantic information about the context word itself.
In fact, the ability of identifying the important and adverse information in the overall semantic space of the whole review according to the given aspect is the foundation of AALSTM's aspect-aware gate mechanism and the reason why it can generate more effective hidden states tailored for ABSA task.

\subsection{Investigation on the Effect of Aspect Numbers}
\begin{table}[b]
\centering
\fontsize{10}{13}\selectfont
\caption{Detailed statistics of the test sets of the three datasets. \textit{\#Aspect\_Number} denotes the subset in which there are \textit{Aspect\_number} aspects contained in the same review.}
\linespread{2}
\begin{tabular}{l|cccccccc}
\toprule \midrule
Test Set Group        & \#1 & \#2 & \#3 & \#4 & \#5 & \#6 & \#7 & \#13 \\ \hline
ATSA-Restaurant & 284 & 371 & 233 & 124 & 70  & 18  & 7   & 13   \\ \hline
ATSA-Laptop     & 257 & 204 & 101 & 40  & 30  & 6   & -   & -      \\ \hline
ACSA-Restaurant & 578 & 293 & 96  & 6   & -   & -   & -   & -    \\ \bottomrule
\end{tabular}
\label{test-aspnum}
\end{table}

\begin{figure*}[t]
 \centering
 \includegraphics[width = \textwidth]{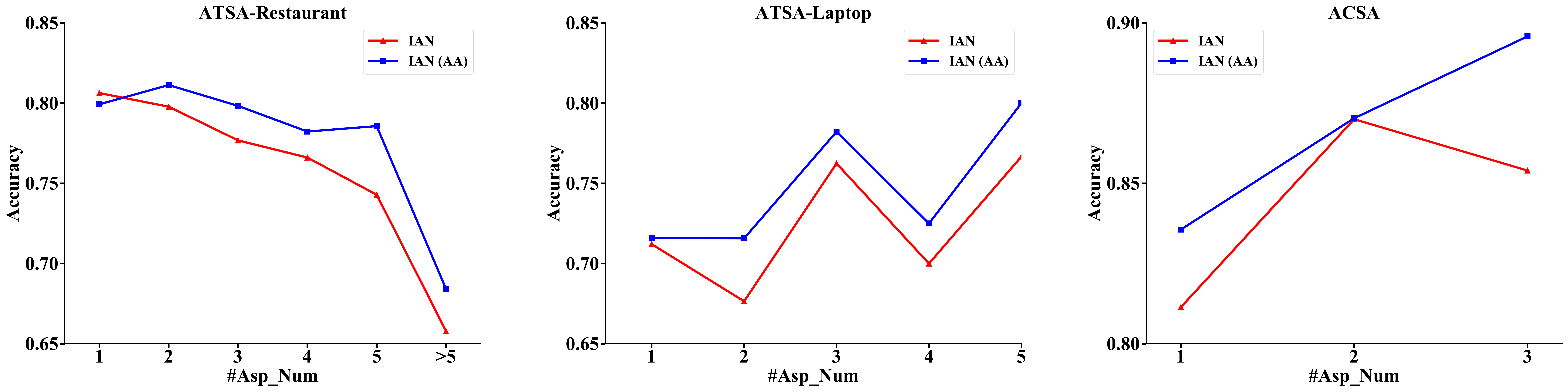}
 \caption{Performance comparison of LSTM-based models on different aspect numbers.}
 \label{chart-asp-ian}
\end{figure*}

\begin{figure*}[t]
 \centering
 \includegraphics[width = \textwidth]{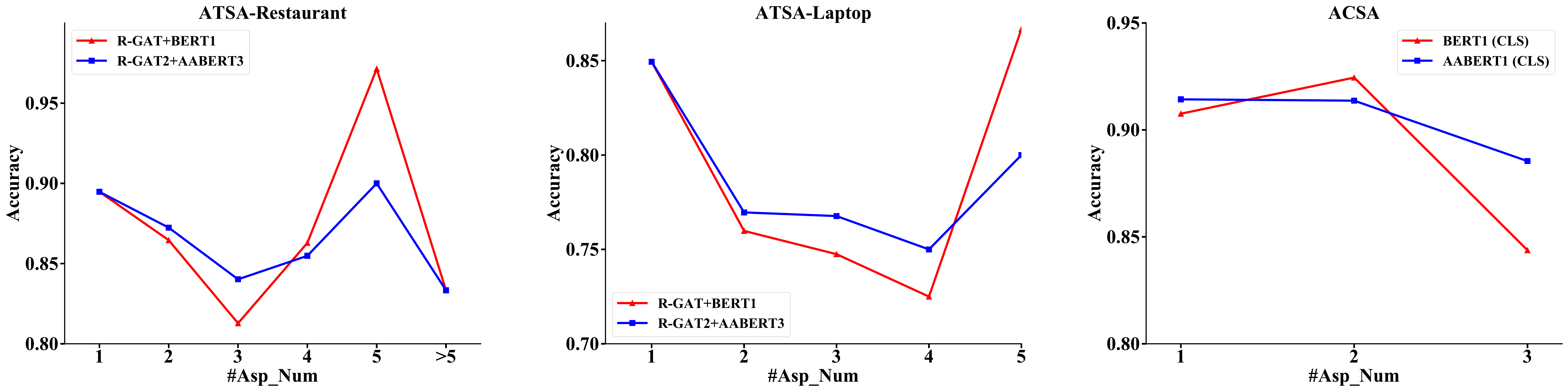}
 \caption{Performance comparison pf BERT-based models on different aspect numbers.}
 \label{chart-asp-gat}
\end{figure*}

In both real-world scenarios and the SemEval 2014 datasets, it is a common phenomenon that multiple aspects are mentioned in the same review.
As revealed in previous work \cite{asgcn}, the adverse interactive affect between different aspects contained in the same review limits ABSA models' performances to a large extent.
The fundamental cause of the multi-aspect problem is that the sentiment indicative information of different aspects is intertwined and interactively affects each other in the context modeling process.
Even if a perfect attention mechanism is designed and learned, this problem can not be resolved.
This is because that when predicting the sentiment of a given aspect, the harmful information for other aspects is contained in the generated hidden states, which can not be changed by the attention mechanism.
In AALSTM, the aspect gate mechanism helps the three LSTM gates only retain the beneficial information and filter out the adverse information regarding the given aspect.
In AABERTs, our designed input format can make BERT model the intra-sentence dependencies between the given aspect and their sentiment trigger words, retaining aspect-related and eliminating aspect-irrelevant information in generated hidden states.
Theoretically, AALSTM and AABERTs can effectively alleviate the multi-aspect adversely interactive affect and improve ABSA model's performance in the multi-aspect situation. 
To prove this, we evaluate the models' performances on different test set groups divided based on the number of aspects existing in one context.
Table \ref{test-aspnum} lists the statistics of the divided test sets. 
To eliminate the result uncertainty, we group the instances of \#6, \#7 and \#13 subsets in the test set of \textit{ATSA-Restaurant} into one subset: \textgreater5, and remove the \#6 subset from the test set of \textit{ATSA-Laptop} as well as \#4 subset from the test set of \textit{ACSA-Restaurant}. The performance comparisons is illustrated in Fig. \ref{chart-asp-ian} and Fig. \ref{chart-asp-gat}.

As illustrated in Fig. \ref{chart-asp-ian}, AALSTM boosts IAN's performance in the multi-aspect situation.
On different multi-aspect number classes, AALSTM is stably superior to vanilla LSTM.
On \textit{ATSA-Restaurant} test set, when \textit{\#Aspect\_Number} rises to 2 from 1 as well as from 4 to 5, IAN and IAN (AA) have dramatically opposite performance trends: IAN's performance drops while IAN (AA)'s performance increases.
On \textit{ATSA-Laptop} test set, when \textit{\#Aspect\_Number} changes to 2 from 1, IAN (AA)'s accuracy nearly remains unchanged, but IAN's performs worse significantly.
Remarkably, on ACSA task, when \textit{\#Aspect\_Number} comes to 3, IAN's performance has a significant decline but IAN (AA)'s performance not only has no decline, it even improves, nearly linearly as \textit{\#Aspect\_Number} increases.
From Fig. \ref{chart-asp-gat}, we can observe that on ATSA task the performance trends of R-GAT+BERT1 and R-GAT2+BERT2 are similar as \textit{\#Aspect\_Number} increases.
However, the performance of R-GAT+BERT1 is fluctuated in the multi-aspect situation, while R-GAT2+BERT2 is more stable and generally performs better.
On ACSA task, we can also observe that AABERT1 (CLS) is more stable than BERT1 (CLS) and obtains better results.

\subsection{Cross Domain Evaluation}
\begin{table}[th]
\centering
\caption{Cross Domain evaluation results on ATSA task. $L\rightarrow R$ denotes models are trained on \textit{Laptop} domain while evaluated on \textit{Restaurant} domain, and vice versa. Top 2 scores are \underline{underlined} and best scores are in \textbf{bold}.}
\fontsize{10}{13}\selectfont
\begin{tabular}{l|cc|cc}\toprule \midrule
\multirow{2}{*}{Models}
           & \multicolumn{2}{c|}{L$\rightarrow$ R} & \multicolumn{2}{c}{R$\rightarrow$ L}   \\ \cline{2-5} \addlinespace[0.5ex]
           & F1            & Acc          & F1         & Acc  \\ \midrule
BERT0 (avg)    & 68.63  &79.44  & 74.47 & 77.88 \\
BERT1 (avg)    & 72.72  &80.84  & 75.52 & 79.04 \\ \midrule
AABERT1 (avg)   & \textbf{\underline{73.13}}  &\underline{80.92}  & 75.41 & 78.84 \\
AABERT2 (avg)   & 72.93  &80.74  & \textbf{\underline{75.69}} & \textbf{\underline{79.19}} \\
AABERT3 (avg)   & \underline{72.96}  &\textbf{\underline{80.99}}  & \underline{75.68} & \underline{79.12} \\

\bottomrule
\end{tabular}
\label{transfer results}
\end{table}
Trained from scratch on massive open-domain general texts, BERT is designed to transfer the learned abundant linguistic knowledge to specific domain of down-stream task \cite{bert}.
We design a set of cross-domain evaluation experiments to test BERT and our proposed AABERTs' abilities of learning general domain-independent sentiment semantic.
Evaluation results are demonstrated in Table \ref{transfer results}.
We can find that AABERTs are more robust for domain transfer.
There are two reasons. 
First, AABERTs can more effectively model the intra-sentence dependencies between aspect and their sentiment trigger words.
Second, domain specificity is mainly characterized by the difference between aspects of different domains.
Without thoroughly separating the context and aspect in the embedding space, the semantic space of context and aspect of AABERTs are more smooth.

%% file: conclusion.tex
\section{Conclusion} \label{conclusion}
In this paper, we discover and define the aspect-agnostic problem which widely exists in the context modeling process of ABSA models.
Then we argue that the semantics of the given aspect should be considered as a new clue out of context when modeling the context.
We propose three streams of aspect-aware context encoders: an aspect-aware LSTM (AALSTM), an aspect-aware GCN (AAGCN), and three aspect-aware BERTs (AABERTs) to generate the aspect-aware hidden states tailored for ABSA task.
Specifically, AALSTM adopts the aspect-aware gate mechanism which dynamically adds the beneficial disturbance from the aspect to the original internal values of the three LSTM gates.
In this way, the given aspect can dynamically help regulate the information flow in LSTM cells along time steps.
As a result, AALSTM can retain the sentiment indicative information of the given aspect in generated hidden states and eliminate the useless information regarding the given aspect.
We augment vanilla GCN with an aspect-aware convolution gate, which can regulate the information flow so as to aggregate the aspect-related information from neighbor nodes to current node and discard the uesless information.
Based on BERT, we flexibly modify the settings of segment embedding and \texttt{[SEP]} token to avoid thoroughly isolating context and the concatenated aspect as two individual sentences.
Compared with BERT, our AABERTs can effectively model the aspect-aware intra-dependencies in the context modeling process.
Experimental evaluations verify that AALSTM, AAGCN and AABERTs significantly outperform their vanilla counterparts on ABSA task, demonstrating their capability of alleviating the aspect-agnostic problem.
Additionally, experimental results show that our AAGCN works well with AALSTM and AABERTs, then our aspect-aware models achieve new state-of-the-art in both LSTM-models and BERT-based models.

To our knowledge, this is the first work that focuses on the shortcomings of context encoders and leverages the aspect as a new clue out of context for context modeling in ABSA task.
In this work, we address ABSA task from a new perspective, and this first investigation paves the way to several aspect-aware extensions of the context encoder based on other architectures, such as Convolutional Neural Network, Memory Network, etc.
Our work also provide insights to other NLP tasks, such as relation classification, which aims to predict the relation between two given entity include in a sentence.
In the scenario of relation classification, generally the sentence encoder is not aware of the entity-pair.
Therefore, the 'entity-pair agnostic' problem, which is a variant of the aspect-agnostic problem we point out in this paper, exists in the task of relation classification.
In the future, we are glad to see more advances in solving the aspect-agnostic problem in ABSA task and and we are keen to explore our aspect aware mechanisms in tackling similar problems in other NLP tasks.

%% file: acknowledge.tex
\section*{Acknowledgements}
This work was supported by Australian Research Council  Grant (DP180100106 and \\ DP200101328).

%% file: jair_aa.bbl
\begin{thebibliography}{}

\bibitem[\protect\BCAY{Bahdanau, Cho,\ \BBA\ Bengio}{Bahdanau
  et~al.}{2015}]{NMT}
Bahdanau, D., Cho, K., \BBA\ Bengio, Y. \BBOP2015\BBCP.
\newblock \BBOQ Neural machine translation by jointly learning to align and
  translate.\BBCQ\
\newblock In Bengio, Y.\BBACOMMA\  \BBA\ LeCun, Y.\BEDS, {\Bem ICLR}.

\bibitem[\protect\BCAY{Chen, Sun, Bing,\ \BBA\ Yang}{Chen
  et~al.}{2017}]{Tencent}
Chen, P., Sun, Z., Bing, L., \BBA\ Yang, W. \BBOP2017\BBCP.
\newblock \BBOQ Recurrent attention network on memory for aspect sentiment
  analysis\BBCQ\
\newblock In {\Bem Empirical Methods in Natural Language Processing}, \BPGS\
  452--461.

\bibitem[\protect\BCAY{Chen, Gan, Cheng, Liu,\ \BBA\ Liu}{Chen
  et~al.}{2020}]{distillbert}
Chen, Y.-C., Gan, Z., Cheng, Y., Liu, J., \BBA\ Liu, J. \BBOP2020\BBCP.
\newblock \BBOQ Distilling knowledge learned in {BERT} for text
  generation\BBCQ\
\newblock In {\Bem Proceedings of the 58th Annual Meeting of the Association
  for Computational Linguistics}, \BPGS\ 7893--7905, Online. Association for
  Computational Linguistics.

\bibitem[\protect\BCAY{Chen\ \BBA\ Qian}{Chen\ \BBA\ Qian}{2019}]{transcap}
Chen, Z.\BBACOMMA\  \BBA\ Qian, T. \BBOP2019\BBCP.
\newblock \BBOQ Transfer capsule network for aspect level sentiment
  classification\BBCQ\
\newblock In {\Bem Proceedings of the 57th Annual Meeting of the Association
  for Computational Linguistics}, \BPGS\ 547--556.

\bibitem[\protect\BCAY{Devamanyu, Soujanya, Prateek, Gangeshwar, Erik,\ \BBA\
  Roger}{Devamanyu et~al.}{2018}]{IAD}
Devamanyu, H., Soujanya, P., Prateek, V., Gangeshwar, K., Erik, C., \BBA\
  Roger, Z. \BBOP2018\BBCP.
\newblock \BBOQ Modeling inter-aspect dependencies for aspect-based sentiment
  analysis\BBCQ\
\newblock In {\Bem Conference of the North American Chapter of the Association
  for Computational Linguistics}, \BPGS\ 266--270.

\bibitem[\protect\BCAY{Devlin, Chang, Lee,\ \BBA\ Toutanova}{Devlin
  et~al.}{2019}]{bert}
Devlin, J., Chang, M.-W., Lee, K., \BBA\ Toutanova, K. \BBOP2019\BBCP.
\newblock \BBOQ {BERT}: Pre-training of deep bidirectional transformers for
  language understanding\BBCQ\
\newblock In {\Bem Proceedings of the 2019 Conference of the North {A}merican
  Chapter of the Association for Computational Linguistics: Human Language
  Technologies, Volume 1 (Long and Short Papers)}, \BPGS\ 4171--4186,
  Minneapolis, Minnesota. Association for Computational Linguistics.

\bibitem[\protect\BCAY{Gao, Feng, Song,\ \BBA\ Wu}{Gao
  et~al.}{2019}]{access2019}
Gao, Z., Feng, A., Song, X., \BBA\ Wu, X. \BBOP2019\BBCP.
\newblock \BBOQ Target-dependent sentiment classification with bert.\BBCQ\
\newblock {\Bem IEEE Access}, {\Bem 7}, 154290--154299.

\bibitem[\protect\BCAY{He, Lee, Ng,\ \BBA\ Dahlmeier}{He et~al.}{2018}]{hrd}
He, R., Lee, W.~S., Ng, H.~T., \BBA\ Dahlmeier, D. \BBOP2018\BBCP.
\newblock \BBOQ Exploiting document knowledge for aspect-level sentiment
  classification\BBCQ\
\newblock In {\Bem Annual Meeting of the Association for Computational
  Linguistics}, \BPGS\ 579--585.

\bibitem[\protect\BCAY{Hochreiter\ \BBA\ Schmidhuber}{Hochreiter\ \BBA\
  Schmidhuber}{1997}]{LSTM}
Hochreiter, S.\BBACOMMA\  \BBA\ Schmidhuber, J. \BBOP1997\BBCP.
\newblock \BBOQ Long short-term memory\BBCQ\
\newblock {\Bem Neural computation}, {\Bem 9\/}(8), 1735--1780.

\bibitem[\protect\BCAY{Huang\ \BBA\ Carley}{Huang\ \BBA\ Carley}{2018}]{pcnn}
Huang, B.\BBACOMMA\  \BBA\ Carley, K. \BBOP2018\BBCP.
\newblock \BBOQ Parameterized convolutional neural networks for aspect level
  sentiment classification\BBCQ\
\newblock In {\Bem Proceedings of the 2018 Conference on Empirical Methods in
  Natural Language Processing}, \BPGS\ 1091--1096, Brussels, Belgium.

\bibitem[\protect\BCAY{Huang\ \BBA\ Carley}{Huang\ \BBA\
  Carley}{2019}]{graphatt}
Huang, B.\BBACOMMA\  \BBA\ Carley, K.~M. \BBOP2019\BBCP.
\newblock \BBOQ Syntax-aware aspect level sentiment classification with graph
  attention networks\BBCQ\
\newblock In {\Bem Proceedings of the 2019 Conference on Empirical Methods in
  Natural Language Processing and the 9th International Joint Conference on
  Natural Language Processing}, \BPG\ 5469–5477.

\bibitem[\protect\BCAY{Joshi, Chen, Liu, Weld, Zettlemoyer,\ \BBA\ Levy}{Joshi
  et~al.}{2020}]{spanbert}
Joshi, M., Chen, D., Liu, Y., Weld, D.~S., Zettlemoyer, L., \BBA\ Levy, O.
  \BBOP2020\BBCP.
\newblock \BBOQ {S}pan{BERT}: Improving pre-training by representing and
  predicting spans\BBCQ\
\newblock {\Bem Transactions of the Association for Computational Linguistics},
  {\Bem 8}, 64--77.

\bibitem[\protect\BCAY{Karimi, Rossi,\ \BBA\ Prati}{Karimi et~al.}{2021}]{ICPR}
Karimi, A., Rossi, L., \BBA\ Prati, A. \BBOP2021\BBCP.
\newblock \BBOQ Adversarial training for aspect-based sentiment analysis with
  bert\BBCQ\
\newblock In {\Bem 2020 25th International Conference on Pattern Recognition
  (ICPR)}, \BPGS\ 8797--8803.

\bibitem[\protect\BCAY{Kim}{Kim}{2014}]{textcnn}
Kim, Y. \BBOP2014\BBCP.
\newblock \BBOQ Convolutional neural networks for sentence classification\BBCQ\
\newblock In {\Bem Proceedings of the 2014 Conference on Empirical Methods in
  Natural Language Processing ({EMNLP})}, \BPGS\ 1746--1751.

\bibitem[\protect\BCAY{Kingma\ \BBA\ Ba}{Kingma\ \BBA\ Ba}{2015}]{Adam}
Kingma, D.~P.\BBACOMMA\  \BBA\ Ba, J. \BBOP2015\BBCP.
\newblock \BBOQ Adam: A method for stochastic optimization\BBCQ\
\newblock In {\Bem ICLR (Poster)}.

\bibitem[\protect\BCAY{Kiritchenko, Zhu, Cherrt,\ \BBA\ and}{Kiritchenko
  et~al.}{2014}]{NRC2}
Kiritchenko, S., Zhu, X., Cherrt, C., \BBA\ and, S.~M. \BBOP2014\BBCP.
\newblock \BBOQ Nrc-canada-2014: Detecting aspects and sentiment in customer
  reviews\BBCQ\
\newblock In {\Bem International Workshop on Semantic Evaluation}, \BPGS\
  437--442.

\bibitem[\protect\BCAY{LeCun, Bengio,\ \BBA\ Hinton}{LeCun
  et~al.}{2015}]{dl-nature}
LeCun, Y., Bengio, Y., \BBA\ Hinton, G. \BBOP2015\BBCP.
\newblock \BBOQ Deep learning\BBCQ\
\newblock {\Bem Nature}, {\Bem 521\/}(7553), 436--444.

\bibitem[\protect\BCAY{Li, Ma, Guo, Xue,\ \BBA\ Qiu}{Li
  et~al.}{2020}]{bert-attack}
Li, L., Ma, R., Guo, Q., Xue, X., \BBA\ Qiu, X. \BBOP2020\BBCP.
\newblock \BBOQ {BERT}-{ATTACK}: Adversarial attack against {BERT} using
  {BERT}\BBCQ\
\newblock In {\Bem Proceedings of the 2020 Conference on Empirical Methods in
  Natural Language Processing (EMNLP)}, \BPGS\ 6193--6202, Online. Association
  for Computational Linguistics.

\bibitem[\protect\BCAY{Liu\ \BBA\ Zhang}{Liu\ \BBA\ Zhang}{2017}]{zhangyue}
Liu, J.\BBACOMMA\  \BBA\ Zhang, Y. \BBOP2017\BBCP.
\newblock \BBOQ Attention modeling for targeted sentiment\BBCQ\
\newblock In {\Bem Proceedings of the 15th Conference of the European Chapter
  of the Association for Computational Linguistics}, \BPG\ 572–577.

\bibitem[\protect\BCAY{Liu, Ott, Goyal, Du, Joshi, Chen, Levy, Lewis,
  Zettlemoyer,\ \BBA\ Stoyanov}{Liu et~al.}{2019}]{roberta}
Liu, Y., Ott, M., Goyal, N., Du, J., Joshi, M., Chen, D., Levy, O., Lewis, M.,
  Zettlemoyer, L., \BBA\ Stoyanov, V. \BBOP2019\BBCP.
\newblock \BBOQ Roberta: A robustly optimized bert pretraining approach\BBCQ.
\newblock cite arxiv:1907.11692.

\bibitem[\protect\BCAY{Ma, Li, Zhang,\ \BBA\ Wang}{Ma et~al.}{2017}]{IAN}
Ma, D., Li, S., Zhang, X., \BBA\ Wang, H. \BBOP2017\BBCP.
\newblock \BBOQ Interactive attention networks for aspect-level sentiment
  classification\BBCQ\
\newblock In {\Bem International Joint Conference on Artificial Intelligence},
  \BPGS\ 4068--4074.

\bibitem[\protect\BCAY{Pennington, Socher,\ \BBA\ Manning}{Pennington
  et~al.}{2014}]{Glove}
Pennington, J., Socher, R., \BBA\ Manning, C.~D. \BBOP2014\BBCP.
\newblock \BBOQ Glove: Global vectors for word representation\BBCQ\
\newblock In {\Bem Empirical Methods in Natural Language Processing (EMNLP)},
  \BPGS\ 1532--1543.

\bibitem[\protect\BCAY{Pontiki, Galanis, Pavlopoulos, Papageorgiou,
  Androutsopoulos,\ \BBA\ Manandhar}{Pontiki et~al.}{2014}]{Semeval2014}
Pontiki, M., Galanis, D., Pavlopoulos, J., Papageorgiou, H., Androutsopoulos,
  I., \BBA\ Manandhar, S. \BBOP2014\BBCP.
\newblock \BBOQ Semeval-2014 task 4: Aspect based sentiment analysis\BBCQ\
\newblock In {\Bem International Workshop on Semantic Evaluation}, \BPGS\
  27--35.

\bibitem[\protect\BCAY{Reimers\ \BBA\ Gurevych}{Reimers\ \BBA\
  Gurevych}{2019}]{s-bert}
Reimers, N.\BBACOMMA\  \BBA\ Gurevych, I. \BBOP2019\BBCP.
\newblock \BBOQ Sentence-{BERT}: Sentence embeddings using {S}iamese
  {BERT}-networks\BBCQ\
\newblock In {\Bem Proceedings of the 2019 Conference on Empirical Methods in
  Natural Language Processing and the 9th International Joint Conference on
  Natural Language Processing (EMNLP-IJCNLP)}, \BPGS\ 3982--3992, Hong Kong,
  China. Association for Computational Linguistics.

\bibitem[\protect\BCAY{Sabour, Frosst,\ \BBA\ Hinton}{Sabour
  et~al.}{2017}]{capsule}
Sabour, S., Frosst, N., \BBA\ Hinton, G.~E. \BBOP2017\BBCP.
\newblock \BBOQ Dynamic routing between capsules\BBCQ\
\newblock In {\Bem Advances in Neural Information Processing Systems 30},
  \BPGS\ 3856--3866.

\bibitem[\protect\BCAY{Scarselli, Gori, Tsoi, Hagenbuchner,\ \BBA\
  Monfardini}{Scarselli et~al.}{2009}]{gnn}
Scarselli, F., Gori, M., Tsoi, A.~C., Hagenbuchner, M., \BBA\ Monfardini, G.
  \BBOP2009\BBCP.
\newblock \BBOQ {The Graph Neural Network Model}\BBCQ\
\newblock {\Bem IEEE Transactions on Neural Networks (TNN)}, {\Bem 20\/}(1),
  61--80.

\bibitem[\protect\BCAY{Song, Wang, Jiang, Liu,\ \BBA\ Rao}{Song
  et~al.}{2019}]{aen-bert}
Song, Y., Wang, J., Jiang, T., Liu, Z., \BBA\ Rao, Y. \BBOP2019\BBCP.
\newblock \BBOQ Attentional encoder network for targeted sentiment
  classification.\BBCQ\
\newblock {\Bem CoRR}, {\Bem abs/1902.09314}.

\bibitem[\protect\BCAY{Sukhbaatar, szlam, Weston,\ \BBA\ Fergus}{Sukhbaatar
  et~al.}{2015}]{end2endMN}
Sukhbaatar, S., szlam, a., Weston, J., \BBA\ Fergus, R. \BBOP2015\BBCP.
\newblock \BBOQ End-to-end memory networks\BBCQ\
\newblock In {\Bem Advances in Neural Information Processing Systems 28},
  \BPGS\ 2440--2448.

\bibitem[\protect\BCAY{Sun, Huang,\ \BBA\ Qiu}{Sun et~al.}{2019}]{bert-qa}
Sun, C., Huang, L., \BBA\ Qiu, X. \BBOP2019\BBCP.
\newblock \BBOQ Utilizing {BERT} for aspect-based sentiment analysis via
  constructing auxiliary sentence\BBCQ\
\newblock In {\Bem Proceedings of the 2019 Conference of the North {A}merican
  Chapter of the Association for Computational Linguistics: Human Language
  Technologies, Volume 1 (Long and Short Papers)}, \BPGS\ 380--385,
  Minneapolis, Minnesota. Association for Computational Linguistics.

\bibitem[\protect\BCAY{Tang, Qin, Feng,\ \BBA\ Liu}{Tang
  et~al.}{2016a}]{TDLSTM}
Tang, D., Qin, B., Feng, X., \BBA\ Liu, T. \BBOP2016a\BBCP.
\newblock \BBOQ Effective lstms for target-dependent sentiment
  classification\BBCQ\
\newblock In {\Bem International Conference on Computational Linguistics},
  \BPGS\ 3298--3307.

\bibitem[\protect\BCAY{Tang, Qin,\ \BBA\ Liu}{Tang et~al.}{2016b}]{DMN}
Tang, D., Qin, B., \BBA\ Liu, T. \BBOP2016b\BBCP.
\newblock \BBOQ Aspect level sentiment classification with deep memory
  network\BBCQ\
\newblock In {\Bem Empirical Methods in Natural Language Processing}, \BPGS\
  214--224.

\bibitem[\protect\BCAY{Tang, Ji, Li,\ \BBA\ Zhou}{Tang et~al.}{2020}]{DGEDT}
Tang, H., Ji, D., Li, C., \BBA\ Zhou, Q. \BBOP2020\BBCP.
\newblock \BBOQ Dependency graph enhanced dual-transformer structure for
  aspect-based sentiment classification\BBCQ\
\newblock In {\Bem Proceedings of the 58th Annual Meeting of the Association
  for Computational Linguistics}, \BPGS\ 6578--6588, Online. Association for
  Computational Linguistics.

\bibitem[\protect\BCAY{Tay, Anh,\ \BBA\ Cheung}{Tay et~al.}{2017}]{Dm}
Tay, Y., Anh, T.~L., \BBA\ Cheung, H.~S. \BBOP2017\BBCP.
\newblock \BBOQ Dyadic memory networks for aspect-based sentiment
  analysis\BBCQ\
\newblock In {\Bem ACM on Conference on Information and Knowledge Management,
  CIKM}, \BPGS\ 107--116.

\bibitem[\protect\BCAY{Tay, Tuan,\ \BBA\ Hui}{Tay et~al.}{2018}]{Fusion}
Tay, Y., Tuan, L.~A., \BBA\ Hui, S.~C. \BBOP2018\BBCP.
\newblock \BBOQ Learning to attend via word-aspect associative fusion for
  aspect-based sentiment analysis\BBCQ\
\newblock In {\Bem AAAI Conference on Artificial Intelligence}, \BPGS\
  5956--5963.

\bibitem[\protect\BCAY{Thomas N.~Kipf}{Thomas N.~Kipf}{2017}]{gcn}
Thomas N.~Kipf, M.~W. \BBOP2017\BBCP.
\newblock \BBOQ Semi-supervised classification with graph convolutional
  networks\BBCQ\
\newblock In {\Bem International Conference on Learning Representations}.

\bibitem[\protect\BCAY{Tian, Chen,\ \BBA\ Song}{Tian et~al.}{2021a}]{tgcn}
Tian, Y., Chen, G., \BBA\ Song, Y. \BBOP2021a\BBCP.
\newblock \BBOQ Aspect-based sentiment analysis with type-aware graph
  convolutional networks and layer ensemble\BBCQ\
\newblock In {\Bem Proceedings of the 2021 Conference of the North American
  Chapter of the Association for Computational Linguistics: Human Language
  Technologies (NAACL)}, \BPGS\ 2910--2922.

\bibitem[\protect\BCAY{Tian, Chen,\ \BBA\ Song}{Tian et~al.}{2021b}]{kvmn-eacl}
Tian, Y., Chen, G., \BBA\ Song, Y. \BBOP2021b\BBCP.
\newblock \BBOQ Enhancing aspect-level sentiment analysis with word
  dependencies\BBCQ\
\newblock In {\Bem Proceedings of the 16th Conference of the European Chapter
  of the Association for Computational Linguistics: Main Volume}, \BPGS\
  3726--3739, Online. Association for Computational Linguistics.

\bibitem[\protect\BCAY{Vaswani, Shazeer, Parmar, Uszkoreit, Jones, Gomez,
  Kaiser,\ \BBA\ Polosukhin}{Vaswani et~al.}{2017}]{transformer}
Vaswani, A., Shazeer, N., Parmar, N., Uszkoreit, J., Jones, L., Gomez, A.~N.,
  Kaiser, L.~u., \BBA\ Polosukhin, I. \BBOP2017\BBCP.
\newblock \BBOQ Attention is all you need\BBCQ\
\newblock In {\Bem Advances in Neural Information Processing Systems 30},
  \BPGS\ 5998--6008.

\bibitem[\protect\BCAY{Veličković, Cucurull, Casanova, Romero, Liò,\ \BBA\
  Bengio}{Veličković et~al.}{2018}]{gat}
Veličković, P., Cucurull, G., Casanova, A., Romero, A., Liò, P., \BBA\
  Bengio, Y. \BBOP2018\BBCP.
\newblock \BBOQ Graph attention networks\BBCQ\
\newblock In {\Bem International Conference on Learning Representations}.

\bibitem[\protect\BCAY{Wagner, Arora, Cortes, Barman, Bogdanova, Foster,\ \BBA\
  Tounsi}{Wagner et~al.}{2014}]{DCU}
Wagner, J., Arora, P., Cortes, S., Barman, U., Bogdanova, D., Foster, J., \BBA\
  Tounsi, L. \BBOP2014\BBCP.
\newblock \BBOQ Dcu: Aspect-based polarity classification for semeval task
  4\BBCQ\
\newblock In {\Bem International Workshop on Semantic Evaluation}, \BPGS\
  223--229.

\bibitem[\protect\BCAY{Wang, Shang, Lioma, Jiang, Yang, Liu,\ \BBA\
  Simonsen}{Wang et~al.}{2021}]{posembed}
Wang, B., Shang, L., Lioma, C., Jiang, X., Yang, H., Liu, Q., \BBA\ Simonsen,
  J.~G. \BBOP2021\BBCP.
\newblock \BBOQ On position embeddings in bert\BBCQ\
\newblock In {\Bem International Conference on Learning Representations}.

\bibitem[\protect\BCAY{Wang, Shen, Yang, Quan,\ \BBA\ Wang}{Wang
  et~al.}{2020}]{RGAT}
Wang, K., Shen, W., Yang, Y., Quan, X., \BBA\ Wang, R. \BBOP2020\BBCP.
\newblock \BBOQ Relational graph attention network for aspect-based sentiment
  analysis\BBCQ\
\newblock In {\Bem Proceedings of the 58th Annual Meeting of the Association
  for Computational Linguistics}, \BPGS\ 3229--3238, Online. Association for
  Computational Linguistics.

\bibitem[\protect\BCAY{Wang, Mazumder, Liu, Zhou,\ \BBA\ Chang}{Wang
  et~al.}{2018}]{tsmn}
Wang, S., Mazumder, S., Liu, B., Zhou, M., \BBA\ Chang, Y. \BBOP2018\BBCP.
\newblock \BBOQ Target-sensitive memory networks for aspect sentiment
  classification\BBCQ\
\newblock In {\Bem Annual Meeting of the Association for Computational
  Linguistics,}, \BPGS\ 957--967.

\bibitem[\protect\BCAY{Wang, Huang,\ \BBA\ Zhao}{Wang et~al.}{2016}]{ATAE}
Wang, Y., Huang, M., \BBA\ Zhao, L. \BBOP2016\BBCP.
\newblock \BBOQ Attention-based lstm for aspect-level sentiment
  classification\BBCQ\
\newblock In {\Bem Empirical Methods in Natural Language Processing}, \BPGS\
  606--615.

\bibitem[\protect\BCAY{Wang, Sun, Han, Liu,\ \BBA\ Zhu}{Wang
  et~al.}{2018}]{sacap}
Wang, Y., Sun, A., Han, J., Liu, Y., \BBA\ Zhu, X. \BBOP2018\BBCP.
\newblock \BBOQ Sentiment analysis by capsules\BBCQ\
\newblock In {\Bem Proceedings of the 2018 World Wide Web Conference}, \BPG\
  1165–1174.

\bibitem[\protect\BCAY{Wang, Sun, Huang,\ \BBA\ Zhu}{Wang
  et~al.}{2019}]{absacap}
Wang, Y., Sun, A., Huang, M., \BBA\ Zhu, X. \BBOP2019\BBCP.
\newblock \BBOQ Aspect-level sentiment analysis using as-capsules\BBCQ\
\newblock In {\Bem The World Wide Web Conference}, \BPG\ 2033–2044.

\bibitem[\protect\BCAY{Xin, Lidong, Wai,\ \BBA\ Bei}{Xin et~al.}{2018}]{Lixin}
Xin, L., Lidong, B., Wai, L., \BBA\ Bei, S. \BBOP2018\BBCP.
\newblock \BBOQ Transformation networks for target-oriented sentiment
  classification\BBCQ\
\newblock In {\Bem Proceedings of the 56th Annual Meeting of the Association
  for Computational Linguistics (Volume 1: Long Papers)}, \BPGS\ 946--956.

\bibitem[\protect\BCAY{Xing, Liao, Song, Wang, Zhang, Wang,\ \BBA\ Huang}{Xing
  et~al.}{2019}]{AA}
Xing, B., Liao, L., Song, D., Wang, J., Zhang, F., Wang, Z., \BBA\ Huang, H.
  \BBOP2019\BBCP.
\newblock \BBOQ Earlier attention? aspect-aware lstm for aspect-based sentiment
  analysis\BBCQ\
\newblock In {\Bem Proceedings of 28th International Joint Conference on
  Artificial Intelligence (IJCAI)}, \BPGS\ 5313--5319.

\bibitem[\protect\BCAY{Xing\ \BBA\ Tsang}{Xing\ \BBA\ Tsang}{2021}]{kagrmn}
Xing, B.\BBACOMMA\  \BBA\ Tsang, I.~W. \BBOP2021\BBCP.
\newblock \BBOQ Understand me, if you refer to aspect knowledge:
  Knowledge-aware gated recurrent memory network\BBCQ\
\newblock {\Bem CoRR}, {\Bem abs/2108.02352}.

\bibitem[\protect\BCAY{Xing\ \BBA\ Tsang}{Xing\ \BBA\ Tsang}{2022}]{dignet}
Xing, B.\BBACOMMA\  \BBA\ Tsang, I.~W. \BBOP2022\BBCP.
\newblock \BBOQ Dignet: Digging clues from local-global interactive graph for
  aspect-level sentiment classification\BBCQ\
\newblock {\Bem CoRR}, {\Bem abs/2201.00989}.

\bibitem[\protect\BCAY{Xu, Liu, Shu,\ \BBA\ Yu}{Xu et~al.}{2019}]{bert_post}
Xu, H., Liu, B., Shu, L., \BBA\ Yu, P.~S. \BBOP2019\BBCP.
\newblock \BBOQ Bert post-training for review reading comprehension and
  aspect-based sentiment analysis.\BBCQ\
\newblock In Burstein, J., Doran, C., \BBA\ Solorio, T.\BEDS, {\Bem NAACL-HLT
  (1)}, \BPGS\ 2324--2335. Association for Computational Linguistics.

\bibitem[\protect\BCAY{Xue\ \BBA\ Li}{Xue\ \BBA\ Li}{2018}]{gcae}
Xue, W.\BBACOMMA\  \BBA\ Li, T. \BBOP2018\BBCP.
\newblock \BBOQ Aspect based sentiment analysis with gated convolutional
  networks\BBCQ\
\newblock In {\Bem Annual Meeting of the Association for Computational
  Linguistics}, \BPGS\ 2514--2523.

\bibitem[\protect\BCAY{Yang, Zeng, Yang, Song,\ \BBA\ Xu}{Yang
  et~al.}{2021}]{chinese-oriented}
Yang, H., Zeng, B., Yang, J., Song, Y., \BBA\ Xu, R. \BBOP2021\BBCP.
\newblock \BBOQ A multi-task learning model for chinese-oriented aspect
  polarity classification and aspect term extraction.\BBCQ\
\newblock {\Bem Neurocomputing}, {\Bem 419}, 344--356.

\bibitem[\protect\BCAY{Yang, Tu, Wang, Xu,\ \BBA\ Chen}{Yang
  et~al.}{2017}]{aaai2017}
Yang, M., Tu, W., Wang, J., Xu, F., \BBA\ Chen, X. \BBOP2017\BBCP.
\newblock \BBOQ Attention based lstm for target dependent sentiment
  classification\BBCQ\
\newblock In {\Bem AAAI Conference on Artificial Intelligence}, \BPG\
  5013–5014.

\bibitem[\protect\BCAY{Zhang, Li, Xu, Leung, Chen,\ \BBA\ Ye}{Zhang
  et~al.}{2020}]{kgcap}
Zhang, B., Li, X., Xu, X., Leung, K.-C., Chen, Z., \BBA\ Ye, Y. \BBOP2020\BBCP.
\newblock \BBOQ Knowledge guided capsule attention network for aspect-based
  sentiment analysis.\BBCQ\
\newblock {\Bem IEEE ACM Trans. Audio Speech Lang. Process.}, {\Bem 28},
  2538--2551.

\bibitem[\protect\BCAY{Zhang, Li,\ \BBA\ Song}{Zhang et~al.}{2019}]{asgcn}
Zhang, C., Li, Q., \BBA\ Song, D. \BBOP2019\BBCP.
\newblock \BBOQ Aspect-based sentiment classification with aspect-specific
  graph convolutional networks\BBCQ\
\newblock In {\Bem Proceedings of the 2019 Conference on Empirical Methods in
  Natural Language Processing and the 9th International Joint Conference on
  Natural Language Processing}, \BPG\ 4568–4578.

\bibitem[\protect\BCAY{Zhang\ \BBA\ Qian}{Zhang\ \BBA\ Qian}{2020}]{bigcn}
Zhang, M.\BBACOMMA\  \BBA\ Qian, T. \BBOP2020\BBCP.
\newblock \BBOQ Convolution over hierarchical syntactic and lexical graphs for
  aspect level sentiment analysis\BBCQ\
\newblock In {\Bem Proceedings of the 2020 Conference on Empirical Methods in
  Natural Language Processing (EMNLP)}, \BPGS\ 3540--3549, Online. Association
  for Computational Linguistics.

\bibitem[\protect\BCAY{Zhang, Han, Liu, Jiang, Sun,\ \BBA\ Liu}{Zhang
  et~al.}{2019}]{ernie}
Zhang, Z., Han, X., Liu, Z., Jiang, X., Sun, M., \BBA\ Liu, Q. \BBOP2019\BBCP.
\newblock \BBOQ {ERNIE}: Enhanced language representation with informative
  entities\BBCQ\
\newblock In {\Bem Proceedings of the 57th Annual Meeting of the Association
  for Computational Linguistics}, \BPGS\ 1441--1451, Florence, Italy.
  Association for Computational Linguistics.

\bibitem[\protect\BCAY{Zheng, Zhang, Mensah,\ \BBA\ Mao}{Zheng
  et~al.}{2020}]{walk}
Zheng, Y., Zhang, R., Mensah, S., \BBA\ Mao, Y. \BBOP2020\BBCP.
\newblock \BBOQ Replicate, walk, and stop on syntax: an effective neural
  network model for aspect-level sentiment classification\BBCQ\
\newblock In {\Bem AAAI}.

\end{thebibliography}
